\documentclass[nonacm=true, review=false, anonymous = false]{acmart}

\AtBeginDocument{%
  \providecommand\BibTeX{{%
    \normalfont B\kern-0.5em{\scshape i\kern-0.25em b}\kern-0.8em\TeX}}}

\usepackage{amsmath}
\usepackage{xspace}
\usepackage{dsfont}
\usepackage{algorithm}
\usepackage{algpseudocode}
\usepackage{mathtools}
\usepackage{dirtree}
\usepackage{multirow}
\usepackage{array}
\usepackage{cellspace}
\usepackage{url}
\usepackage{hyperref}
\usepackage{fancyvrb}
\usepackage{subcaption}

\newcolumntype{M}[1]{>{\centering\arraybackslash}m{#1}}

\DefineVerbatimEnvironment{CodeInput}{Verbatim}{}

\DeclareMathOperator{\mean}{mean}

\begin{document}
\title{MA-BBOB: A Problem Generator for Black-Box Optimization Using Affine Combinations and Shifts}

\author{Diederick Vermetten}
\affiliation{%
  \institution{LIACS, Leiden University}
  \city{Leiden}
  \country{The Netherlands}}
\email{d.l.vermetten@liacs.leidenuniv.nl}
\orcid{0000-0003-3040-7162}

\author{Furong Ye}
\affiliation{%
  \institution{Institute of Software, Chinese Academy of Science}
  \city{Beijing}
  \country{China}}
\email{f.ye@ios.ac.cn}
\orcid{0000-0002-8707-4189}

\author{Thomas B{\"a}ck}
\affiliation{%
  \institution{LIACS, Leiden University}
  \city{Leiden}
  \country{The Netherlands}}
\email{t.h.w.baeck@liacs.leidenuniv.nl}
\orcid{0000-0001-6768-1478}

\author{Carola Doerr}
\affiliation{%
  \institution{Sorbonne Université, CNRS, LIP6}
  \city{Paris}
  \country{France}}
\email{carola.doerr@lip6.fr}
\orcid{0000-0002-4981-3227}

\renewcommand{\shortauthors}{Vermetten et al.}

\begin{abstract}
Choosing a set of benchmark problems is often a key component of any empirical evaluation of iterative optimization heuristics. In continuous, single-objective optimization, several sets of problems have become widespread, including the well-established BBOB suite. While this suite is designed to enable rigorous benchmarking, it is also commonly used for testing methods such as algorithm selection, which the suite was never designed around. 

We present the MA-BBOB function generator, which uses the BBOB suite as component functions in an affine combination. In this work, we describe the full procedure to create these affine combinations and highlight the trade-offs of several design decisions, specifically the choice to place the optimum uniformly at random in the domain. We then illustrate how this generator can be used to gain more low-level insight into the function landscapes through the use of exploratory landscape analysis. 

Finally, we show a potential use-case of MA-BBOB in generating a wide set of training and testing data for algorithm selectors. Using this setup, we show that the basic scheme of using a set of landscape features to predict the best algorithm does not lead to optimal results, and that an algorithm selector trained purely on the BBOB functions generalizes poorly to the affine combinations. 
\end{abstract}

\ccsdesc[500]{Theory of computation~Random search heuristics}
\ccsdesc[500]{Theory of computation~Design and analysis of algorithms}
\ccsdesc[300]{Theory of computation~Bio-inspired optimization}

\keywords{evolutionary computation, black-box optimization, benchmarking, algorithm selection}

\maketitle

\section{Introduction}

Benchmarking is an essential step in the analysis and comparison of iterative optimization heuristics. By studying an algorithm's behavior on sets of known problems, insights into their working principles can be gained, allowing researchers to improve their algorithms. Because of this interaction between benchmark problems and algorithm design, the availability of benchmark problems inherently influences the kinds of research questions being asked and the ways in which they are answered. 

Since designing effective benchmark suites is a challenging undertaking, rigorously designed suites can become widely adopted in the community. Easy access to these established benchmark suites then leads to these suites being used more often, for a wider variety of studies whose aims might not necessarily align with the design choices made in the construction of the benchmark. This can lead to biases in the results when algorithms exploit properties of the benchmark, e.g., center-bias in many commonly used problems~\cite{kudela2022critical}. 

In continuous, single-objective optimization, the Black Box Optimization Benchmark (BBOB)~\cite{bbobfunctions} is one of the most well-established suites, partly thanks to its integration into the popular COCO framework~\cite{hansen2020coco}. The BBOB suite consists of 24 problems, which each represent several high-level landscape characteristics which challenge optimization algorithms in different ways. For each of these problems, instances can be created through a set of transformations, allowing researchers to test several invariances of their algorithm. Because of its popularity, studies into the specifics of the BBOB suite are numerous~\cite{munoz2015algorithm, evostar_bbob_instance, ela2_munoz2022}, as are studies into the instance landscape which is not covered by BBOB~\cite{NewBBOB-ISA-MunozS20, long2023challenges, tian2020recommender}.   

One key technique used in the analysis of benchmark functions is Exploratory Landscape Analysis (ELA)~\cite{mersmann2011exploratory}. ELA makes use of a set of randomly sampled points in the domain to calculate sets of low-level landscape features, which in aggregate aim to characterize the high-level properties of the landscape. Studies of ELA on BBOB have identified that it is relatively straightforward to detect differences between all 24 problems~\cite{renau2021towards}, while different instances of the same problem generally have comparable feature representations (although this is not the case for all functions~\cite{evostar_bbob_instance}).  

The ability of ELA to differentiate between BBOB functions suggests that these features would be a useful representation for the algorithm selection problem as well. Indeed, studies on algorithm selection for continuous optimization heuristics often use BBOB as their benchmark suite~\cite{KerschkeT19, KerschkeHNT19survey, AnjaPPSN2022, bischl2012algorithm, lacroix2019, munoz2015algorithm}. However, the challenge with using BBOB for algorithm selection lies in the evaluation of the results. One method is a leave-one-function-out technique~\cite{nikolikj2023rf}, which uses 23 functions for training and the remaining one for testing. This approach tends to show poor performance since each problem has been designed to represent different high-level challenges for the optimization algorithm. As such, another technique of cross-validation by splitting function instances is commonly used~\cite{KerschkeT19}. However, this is likely to overfit and overestimate the performance of the selector, since the instances of different problems are inherently very similar. Thus, overfitting to biases of the instance design is an often overlooked risk~\cite{AnjaPPSN2022}.

One potential way in which this bias can be reduced is by creating new, larger sets of benchmark problems, or by creating a problem generator (e.g., the W-model in pseudo-Boolean optimization~\cite{w_model}). Such problem generators can then create arbitrarily many benchmark functions, to be used in the common train/test or cross-validation mechanisms from the machine learning community~\cite{pikalov2022parameter}. 

Extending previous work presented at the GECCO~\cite{ABBOB-GECCO} and the AutoML~\cite{automl_mabbob} conferences, we describe in this work the Many-Affine BBOB function generator (MA-BBOB). To construct new functions, we create affine combinations between existing BBOB functions, building on the work of Dietrich and Mersmann~\cite{affinebbob}. Our generator is a generalization of their approach, designed to create unbiased combinations of problems where the contribution of the components can be smoothly varied. We highlight the core design choices made in the construction of MA-BBOB in Section~\ref{sec:affine_process} and illustrate their impact on the types of problems which can be created. 

The parameterization of the MA-BBOB generator allows us to investigate controlled, potentially small differences in functions from both the ELA and algorithm performance perspectives. This is illustrated in Section~\ref{sec:pairwise_exp} by investigating the addition of global structure (sphere function) to all other BBOB problems, as well as transitioning between pairs of BBOB problems. 

Finally, in Section~\ref{sec:alg_sel} we make use of a set of $1\,000$ functions generated with MA-BBOB to illustrate an algorithm selection scenario, and show that the generalization from BBOB to MA-BBOB fails to meet expectations. A comparison of the ELA-based algorithm selection approach with an artificial baseline using the weights of the affine combinations indicates that there is room to improve on the current ELA-based setup, especially when trying to generalize from BBOB to MA-BBOB. 

\section{Related Work}\label{sec:rel_work}
\subsection{The BBOB Problem Suite}

Within the area of continuous optimization, the BBOB suite stands out as the by far most heavily used one for algorithm comparison and benchmarking. As part of the popular COCO benchmarking framework~\cite{hansen2020coco}, the original BBOB suite~\cite{bbobfunctions} consists of 24 single-objective, noiseless functions. To date, hundreds of different optimization algorithms have been tested on the BBOB suite~\cite{10yearsBBOB}. Over the years, the BBOB suite has been integrated into other benchmarking tools -- as a whole, as is the case for BBOBtorch~\cite{BBOBtorch} and IOHexperimenter~\cite{iohexp}, or in parts, as is the case for the Nevergrad platform~\cite{nevergrad}.  

Of particular note is the fact that the BBOB suite allows for the creation of different instances of each function, by choosing between sets of transformations that are applied to a ``base'' version. These instances aim to preserve the global function properties while varying factors such as the location of the global optimum. This is done to prevent a biased algorithm from exploiting specific structures of the function generation, e.g., a rotational bias. While the transformations are designed to keep the high-level problem structure intact, the impact of these transformations on the low-level landscape properties is not as straightforward, as some features can vary significantly between instances~\cite{evostar_bbob_instance}. 

This instance generation procedure can be seen as one of the reasons for the popularity of the BBOB suite since it allows for the analysis of several invariances in optimization algorithms. Additionally, the availability of instances allows for the evaluation of AutoML approaches, e.g., algorithm selection, by enabling a cross-validation approach across instances. This generally leads to better results than cross-validation on a function-level, since the functions themselves have been designed specifically to contain different challenges to the optimizer, so leaving a function out of the training set generally makes the transfer between train and test challenging~\cite{AnjaPPSN2022, nikolikj2023rf}. 

\subsection{Filling the Instance Space}

While using function instances allows the BBOB suite to cover a wider range of problem landscapes than the raw functions alone, there are limits to the types of landscapes which can be created in this way. Instance space analysis is an emerging research direction which investigates the extent to which the existing benchmark functions cover the instance space, and whether they do so evenly with regard to challenges for different optimization algorithms~\cite{smith2023instance}. This analysis allows for studies into different mechanisms for creating new problems, ranging from random function generators~\cite{tian2020recommender} to genetic programming~\cite{long2023challenges, NewBBOB-ISA-MunozS20}. In parallel, questions regarding appropriate choice of representative benchmark functions from large sets of problems are gaining more attention, since benchmarking on thousands of functions is still not computationally feasible~\cite{cenikj2022selector}.

In addition to the function generation methods mentioned above, it has recently been proposed to use affine combinations between pairs of BBOB functions to generate new benchmark functions~\cite{affinebbob}. These combinations have been shown to smoothly fill the space of low-level landscape properties, as measured through a low-dimensional projection of a set of ELA features~\cite{mersmann2011exploratory}. These results have shown that even a relatively simple function creation procedure has the potential to give us new insights into the inner workings of benchmark problems and their low-level characteristics.

Similarly, the ability to combine existing BBOB functions would allow for more extensive investigations into the behavior of algorithm selection methods, particularly those based on ELA features~\cite{KerschkeHNT19survey}, as smooth transitions in the landscape should correspond to smooth transitions in algorithm behavior as well. Our extension of the affine function combination generation mechanism, which is summarized in Section~\ref{sec:affine_process}, aims to increase the variety of available functions while avoiding issues related to scaling differences between component functions~\cite{ABBOB-GECCO, automl_mabbob}.

\section{Many-Affine BBOB}\label{sec:affine_process}

\subsection{Pairwise Affine Combinations}\label{sec:pairwise_def}

\begin{figure*}[th]
    \centering
    \includegraphics[width=0.97\textwidth]{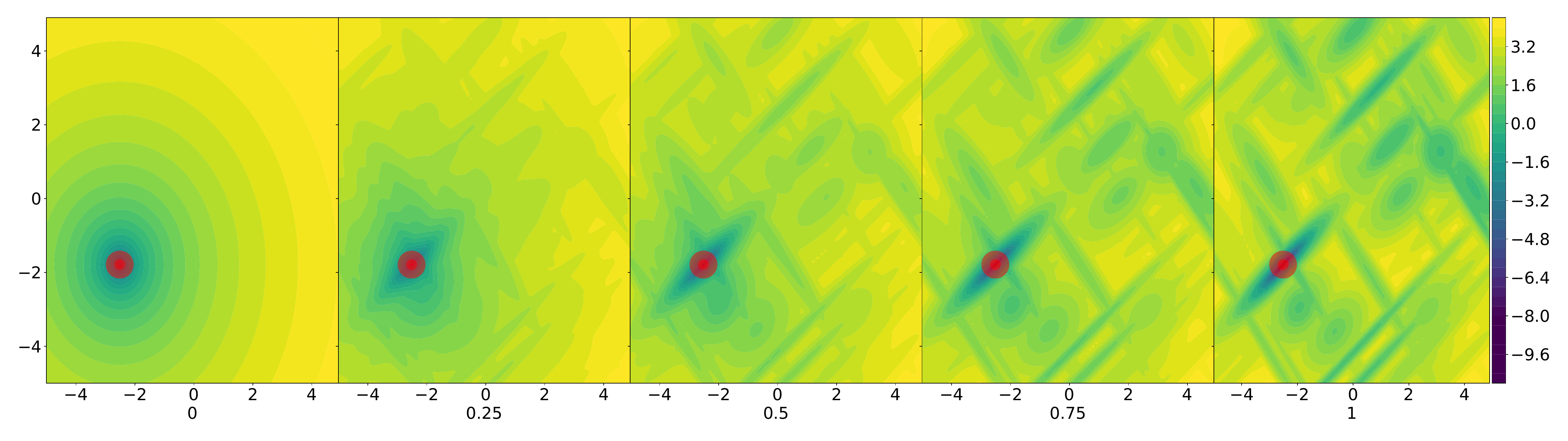}
    \caption{Evolution of the landscape (log-scaled function-values) of the affine combination between F21 ($\alpha=1$) and F1 ($\alpha=0$), instance 1 for both functions, for varying $\alpha$ in 0.25 increments. 
    The red circle highlights the location of the global optimum.}
    \label{fig:frame_f21_f1}
\end{figure*}

To create affine combinations between two BBOB functions, we use a slightly modified version of the procedure proposed in~\cite{affinebbob}. Specifically, we define the combination $C$ as follows:
\begin{align*}
C(F_{1, I_1}, F_{2, I_2}, \alpha)(x) &= 10^X \text{, with} \\
X&=\Big( \alpha \log_{10}\big(F_{1, I_1}(x) - F_{1, I_1}(O_{1, I_1})\big) + (1-\alpha) \log_{10}\big(F_{2, I_2}(x - O_{1, I_1} + O_{2, I_2}) - F_{2, I_2}(O_{2, I_2}) \big)\Big) 
\end{align*}
Here, $F_1$, $I_1$, $F_2$, and $I_2$ are the two base functions and their instance numbers, as defined in BBOB~\cite{bbobfunctions}. $O_{1, I_1}$ and $O_{2, I_2}$ represent the location of the optimum of functions $F_{1, I_1}$ and $F_{2, I_2}$ respectively.  The transformation to $x$ when evaluating $F_{2, I_2}$ is performed to make sure the location of the optimum is at $O_{1, I_1}$. As opposed to the original definition, we subtract the optimal values before aggregating and take a logarithmic mean between the problems. This way, we can use consistent values for $\alpha$ across problems, without having to perform the entropy-based selection performed in~\cite{affinebbob}. It has the additional benefit of ensuring the objective value of the optimal solution is always 0, so the comparison of performance across instances and problems is simplified. In Figure~\ref{fig:frame_f21_f1}, we illustrate the change in the landscape for the combination of F21 and F1, for different values of $\alpha$.

\subsection{Combining Multiple BBOB Functions}

We extend the pairwise affine combinations from Section~\ref{sec:pairwise_def} to create a function generator which uses affine combinations of multiple BBOB functions. In particular, our generator is defined as follows:
\begin{equation*}
    \textit{MA-BBOB}(\Vec{W}, \Vec{I}, \Vec{X}_{\texttt{opt}})(x) = R^{-1}\big(\sum_{i=1}^{24} W_i \cdot R_i\big(F_{i, I_i}(x -\Vec{X}_{\texttt{opt}} + O_{i, I_i}) - F_{i, I_i}(O_{i, I_i})\big)\big)
\end{equation*}
Here, $\Vec{W}$ and $\Vec{I}$ are 24-dimensional vectors containing the weight and instance identifiers respectively, and $\Vec{X}_{\texttt{opt}}$ is the location of the optimum, which we generate uniformly at random in the domain $[-5,5]^d$. Finally, $R_i$ and $R^{-1}$ are rescaling functions, defined in Section~\ref{sec:bbobscaling}. We highlight the motivation behind each of these design choices in the following subsections. 

\subsubsection{Scaling of Function Values}
\label{sec:bbobscaling}

While the geometric weighted average used in Section~\ref{sec:pairwise_def} between component functions reduces the impact of small differences in scale, some BBOB problems vary by orders of magnitude, which can still cause one function to dominate the combined landscape. To address this, we add a rescaling function to the MA-BBOB definition, which transforms the log-precision on each component function into approximately $[0,1]$ before the transformation. This is done by capping the log-precision at $-8$, adding $8$ so the minimum is at $0$ and dividing by a \textit{scale factor} $S_i$. This procedure aims to make the target precision of $10^{2}$ similarly easy to achieve on all component problems. We thus get the following scaling functions:
\begin{align*}
    R_i(x) &= \frac{\max(\log_{10}(x), -8) + 8}{S_i} \\ 
    R^{-1}(x) &= 10^{(10\cdot x) - 8}
\end{align*}

\begin{figure*}
    \centering
    \includegraphics[width=\textwidth]{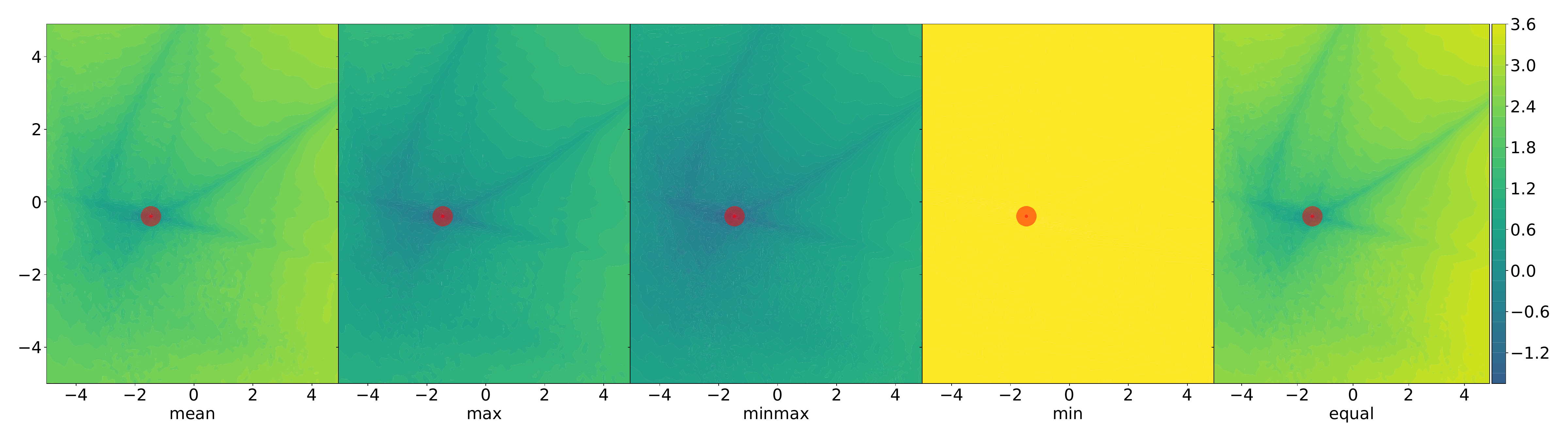}
    \caption{Log-scaled fitness values of an example of a single many-affine function with 5 different ways of scaling. The first 4 are taking the mean, max, $(\max+\min)/2$ and min of $50\,000$ random samples to create the scale factor, while the fifth (`equal') option does not make use of this scaling.}
    \label{fig:scale_factors}
\end{figure*}

\begin{figure}
    \centering
    \includegraphics[width=\textwidth]{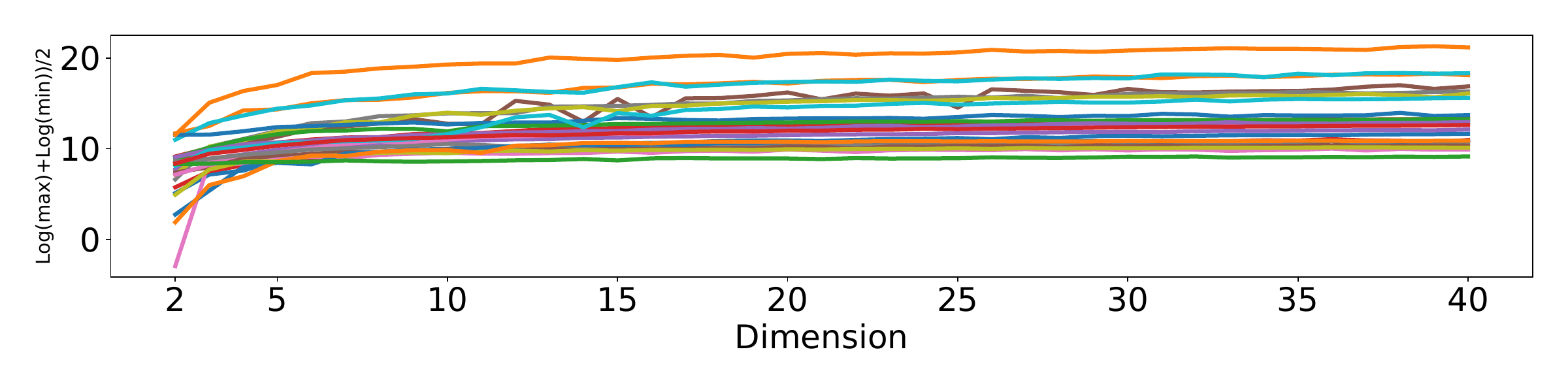}
    \captionsetup{width=\textwidth}

    \caption{Evolution of the log-scaled $(\max+\min)/2$ scaling factor, relative to the problem dimension. The values are based on $50\,000$ samples. Each line corresponds to one of the 24 BBOB functions.}
    \label{fig:scale_factors_per_dim}
\end{figure}

To determine practical scale factors, we collect a set of $50\,000$ random samples and evaluate them. We then aggregate the resulting function values (transformed to log-precision) in several ways: $\min$, $\mean$, $\max$, $(\max+\min)/2$. In Figure~\ref{fig:scale_factors}, we show the differences between these methods for a selected problem in $2d$. Somewhat subjectively, we select the $(\max+\min)/2$ scaling as the technique to use for the MA-BBOB generator. To ensure we don't have to repeat this sampling procedure each time we instantiate the problem in a new dimensionality, we investigate the relation between dimensionality and the chosen scale factor calculation. This is visualized in Figure~\ref{fig:scale_factors_per_dim}, where we see that, with an exception for the smallest dimensions, the values remain quite stable. Because of this, we make use of a static scale factor rather than defining one for each dimension individually. The final factors used are calculated as a rounded median of the values from Figure~\ref{fig:scale_factors_per_dim}, and shown in Table~\ref{tab:scale_factors}.  

\begin{table}[]
    \centering
    \begin{tabular}{lrrrrrrrrrrrr}\toprule{} 
    Function ID   &    1  &    2  &    3  &    4  &    5  &    6  &    7  &    8  &    9  &    10 &    11 &    12 \\

    Scale Factor &  11.0 &  17.5 &  12.3 &  12.6 &  11.5 &  15.3 &  12.1 &  15.3 &  15.2 &  17.4 &  13.4 &  20.4 \\ \hline
    Function ID &    13 &    14 &    15 &    16 &    17 &    18 &    19 &    20 &    21 &    22 &    23  & 24\\
    Scale Factor &  12.9 &  10.4 &  12.3 &  10.3 &   9.8 &  10.6 &  10.0 &  14.7 &  10.7 &  10.8 &   9.0 &  12.1 \\
    \bottomrule
    \end{tabular}
    \caption{Final scale factors used to generate MA-BBOB problems.}
    \label{tab:scale_factors}
\end{table}

\subsubsection{Instance Creation}\label{sec:instances}

\begin{figure*}
    \centering
    \includegraphics[width=\textwidth]{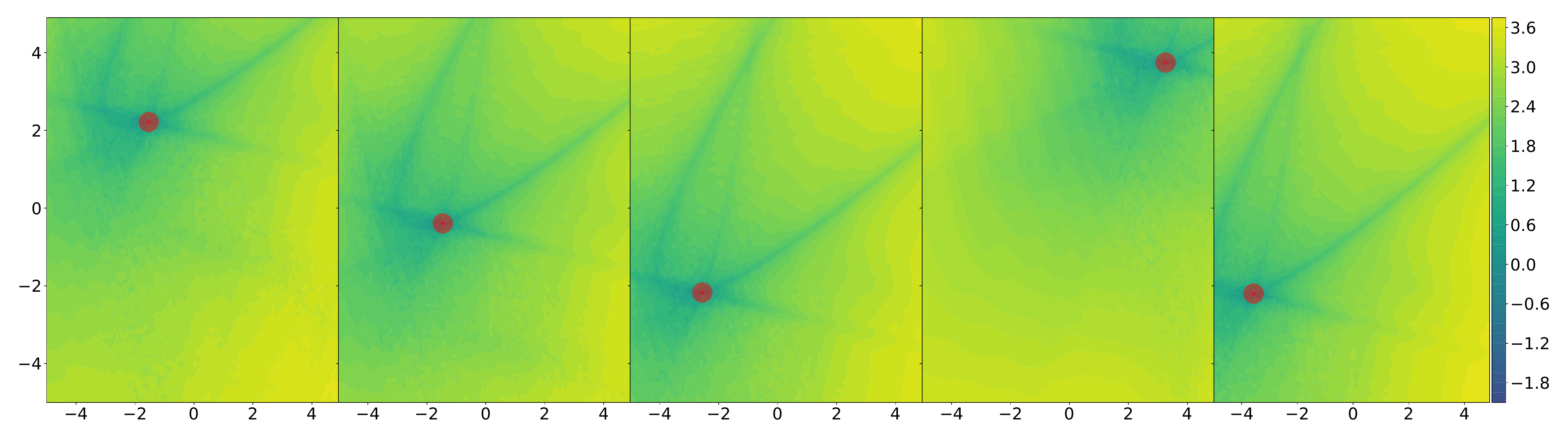}
    \caption{Log-scaled fitness values of an example of a single many-affine function with changed location of optimum.}
    \label{fig:opt_loc}
\end{figure*}

Another design choice we made was to place the optimum of the combined function uniformly in the domain ( $[-5,5]^d$). This differs from the earlier versions used for pairwise combinations of BBOB functions~\cite{affinebbob, ABBOB-GECCO}, where the optimum of one of the component functions was re-used. However, the biases in the original BBOB instance generation procedure would then be transferred into the combinations as well~\cite{evostar_bbob_instance}. 
Since our function generator does not have to guarantee the preservation of global function properties, we take the risk of moving parts of the regions of interest outside the domain to have a less biased location of the global optimum. Figure~\ref{fig:opt_loc} shows how a $2d$-function changes when moving the optimum location. 

\subsubsection{Sampling Random Functions}\label{sub:weights}

To allow for the usage of MA-BBOB as a function generator, we need to create a default setting to generate useful weight-vectors.
This could be done uniformly at random (given a normalization step). However, in this way, the weight for every component is likely to be non-zero, so most functions contribute to the final combination, erasing the possibility of generating unimodal problems since some multimodality will always be included from some of the multimodal component functions.

To address this issue, we adapt the sampling technique to combine fewer component functions on average. Our approach is based on a threshold value to determine which functions contribute to the problem. The procedure for generating weights is thus as follows: (1) Generate initial weights uniformly at random, (2) adapt the threshold to be the minimum of the user-specified threshold and the third-highest weight, (3) this threshold is subtracted from the weights, all negative values are set to 0. 
The second step is to ensure that at least two problems always contribute to the new problem. We decide to set the default value at $T=0.85$, such that on average $3.6$ (i.e., 15\% of 24) problems will have a non-zero weight. 

\subsection{Availability and Reproducibility}

The MA-BBOB generator as described here is made available directly in the IOHexperimenter package~\cite{iohexp} as part of the IOHprofiler project~\cite{IOHprofilerArxiv}. This enables us to use any of the built-in logging and tracking options of IOHexperimenter. In particular, it allows us to store the performance data into a file format which can be directly processed into IOHanalyzer~\cite{IOHanalyzer} for post-processing.  The further experiments in this paper can be reproduced by following the steps outlined in our Zenodo repository~\cite{zenodo_extension}, which also contains the performance data and the ELA-features for all analyzed functions.

\section{Pairwise Affine Combinations}\label{sec:pairwise_exp}

For the first analysis of the MA-BBOB functions, we limit ourselves to the combination of pairs of functions. This allows a more low-level investigation into the transition of both landscape features and algorithm performance. 

\subsection{Setup}\label{sec:setup1}

For the algorithm performance-based analysis, we make use of a portfolio of five algorithms. Of these, three are accessed through the Nevergrad framework~\cite{nevergrad}:
\begin{itemize}
    \item Differential Evolution (DE) ~\cite{de}
    \item Estimation of Multivariate Normal Algorithm (EMNA)~\cite{emna}
    \item Diagonal Covariance Matrix Adaptation Evolution Strategy (dCMA-ES)~\cite{hansen2001self_adaptation_es}
\end{itemize}
The remaining algorithms are two modular algorithm families: modular Differential Evolution~\cite{modDE} (modde) and modular CMA-ES~\cite{nobel_modcma_assessing} (modcma). All algorithms, including the modular ones, use default parameter settings. Each run we perform has a budget of  $2\,000 d$, where $d$ is the dimensionality of the problem. We perform $50$ independent runs per function. For the pairwise function combinations, we stick to the terminology introduced in Section~\ref{sec:pairwise_def} for easier comparison with the previous results in \cite{ABBOB-GECCO}. 

In the remainder of this section, we set $I_2=1$. As such, when discussing the \emph{instance} of a pairwise affine combination $C(F_1, I_1, F_2, I_2, \alpha)$, we are referring to $I_1$. Note that we also introduce the uniform sampling of the optima for these experiments, following the description in Section~\ref{sec:instances}. For our performance measure, we make use of the normalized area over the convergence curve (AOCC), to be maximized. The AOCC is an anytime performance measure, which is equivalent to the area under the cumulative distribution curve (AUC) given infinite targets for the construction of the ECDF. This measure is thus slightly more precise than the AUC, and can easily be computed online. To remain consistent with the performance measures used in our previous work, and analysis of results on BBOB in general, we use $10^2$ and $10^{-8}$ as the bounds for our function values, and perform a log-scaling before calculating the AOCC. We thus calculate the normalized AOCC of a single run as follows:
\begin{equation*}
    \textit{AOCC}(\Vec{y}) = \frac{1}{B} \sum_{i=1}^{B} 1-\frac{ \min(\max((\log_{10}(y_i), -8), 2)  + 8}{10}
\end{equation*}
where $\Vec{y}$ is the sequence of best-so-far function values reached, $B$ is the budget of the run. To obtain the AOCC over multiple runs, we simply take the average. 

For the landscape analysis, we make use of the pFlacco~\cite{pFlacco} package to calculate the ELA features. We use a sample size of $1\,000d$ points, sampled using a Sobol' sequence. We note this large sample size is used to remove some of the inherent variability in the ELA features, even though practical applications usually rely on much smaller budgets. To be consistent with our previous work~\cite{automl_mabbob}, we don't include features which require additional samples and remove all features which lead to NaN-values or remain static for all functions, resulting in a set of $44$ features. 

\subsection{Adding Global Structure}\label{sec:sphere_perf}

For the first set of experiments, we make use of affine combinations where we combine each function with F1: the sphere model (as the $F_2$ function in the combination). As can be seen in Figure~\ref{fig:frame_f21_f1}, adding a sphere model to another function creates an additional global structure that can guide the optimization toward the global optimum. As such, these kinds of combinations might allow us to investigate the influence of an added global structure on the performance of optimization algorithms. While to some extent this can already be investigated by comparing results on the function groups of the original BBOB with different levels of global structure, the affine function combinations allow for a much more fine-grained investigation.

\begin{figure*}
    \centering
    \begin{subfigure}[b]{0.475\textwidth}
        \centering
        \includegraphics[width=\textwidth, trim=0mm 0mm 30mm 8mm,clip]{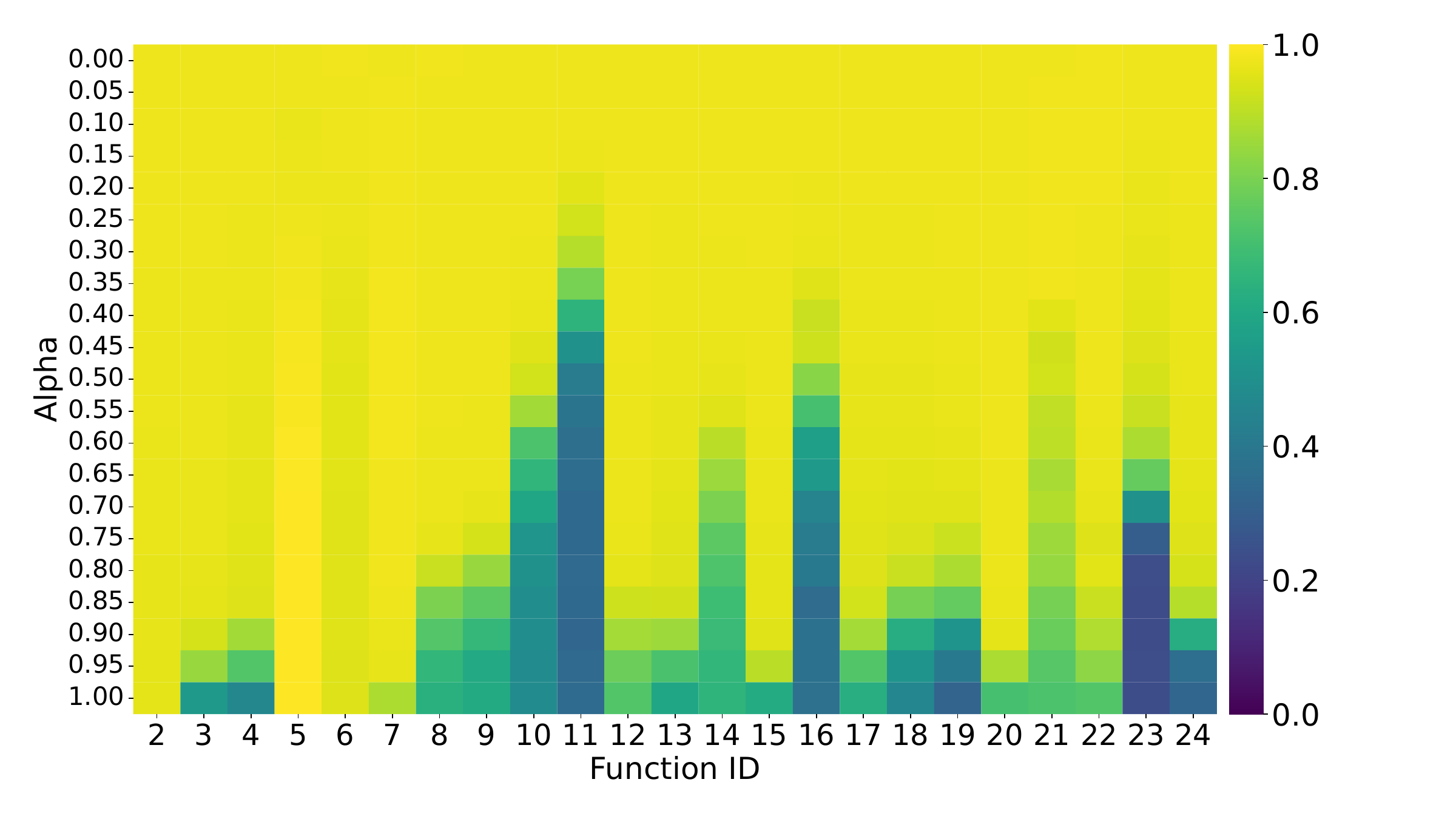}
        \caption{AOCC values for Diagonal CMA-ES.}  
        \label{fig:cma_heatmap_wsphere}
    \end{subfigure}
    \hfill
    \begin{subfigure}[b]{0.475\textwidth}  
        \centering 
        \includegraphics[width=\textwidth, trim=0mm 0mm 30mm 8mm,clip]{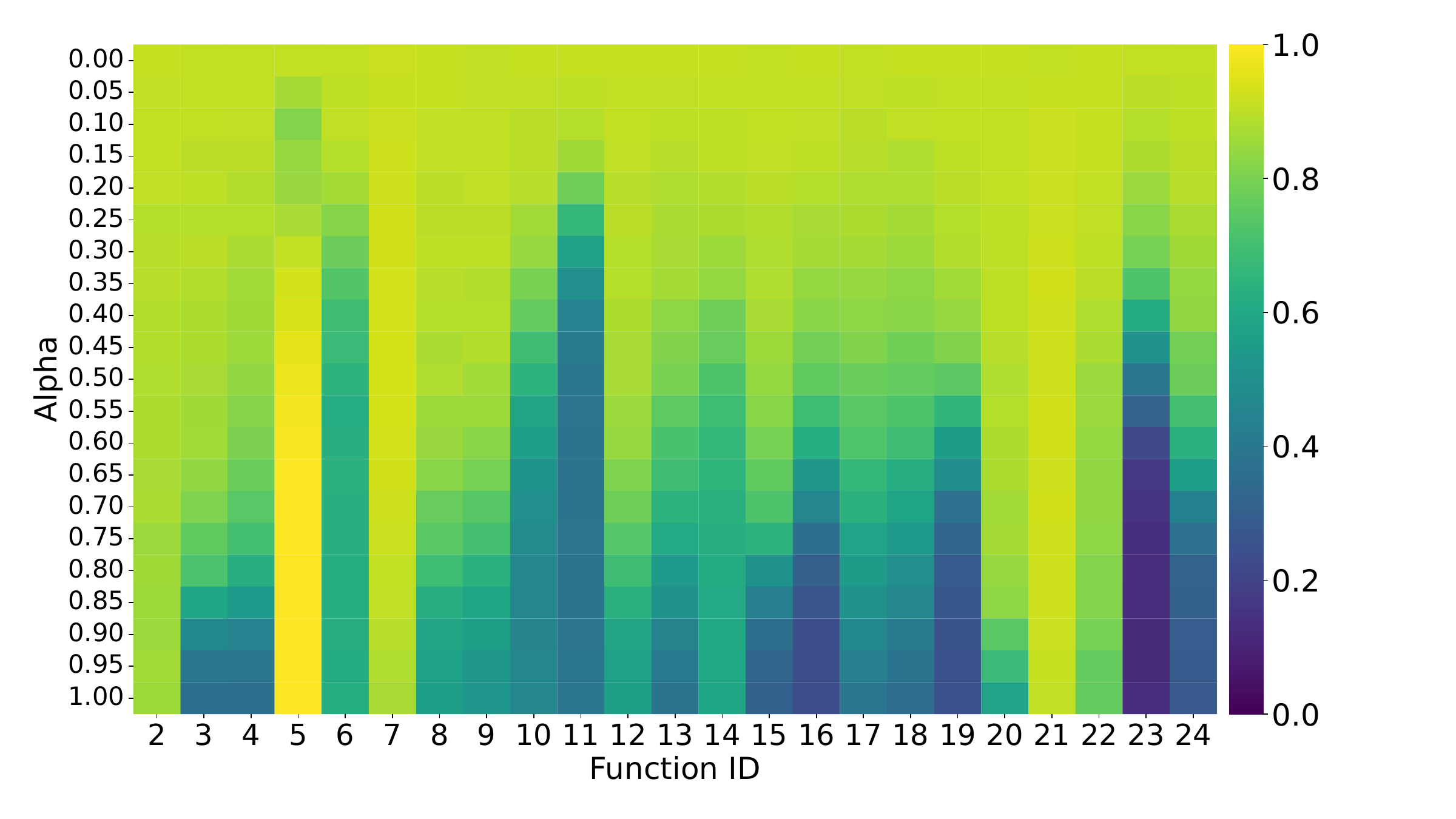}
        \caption{AOCC values for Differential Evolution.}  
        \label{fig:de_heatmap_wsphere}
    \end{subfigure}
    \vskip\baselineskip
    \begin{subfigure}[b]{0.475\textwidth}   
        \centering 
        \includegraphics[width=\textwidth, trim=0mm 0mm 30mm 8mm,clip]{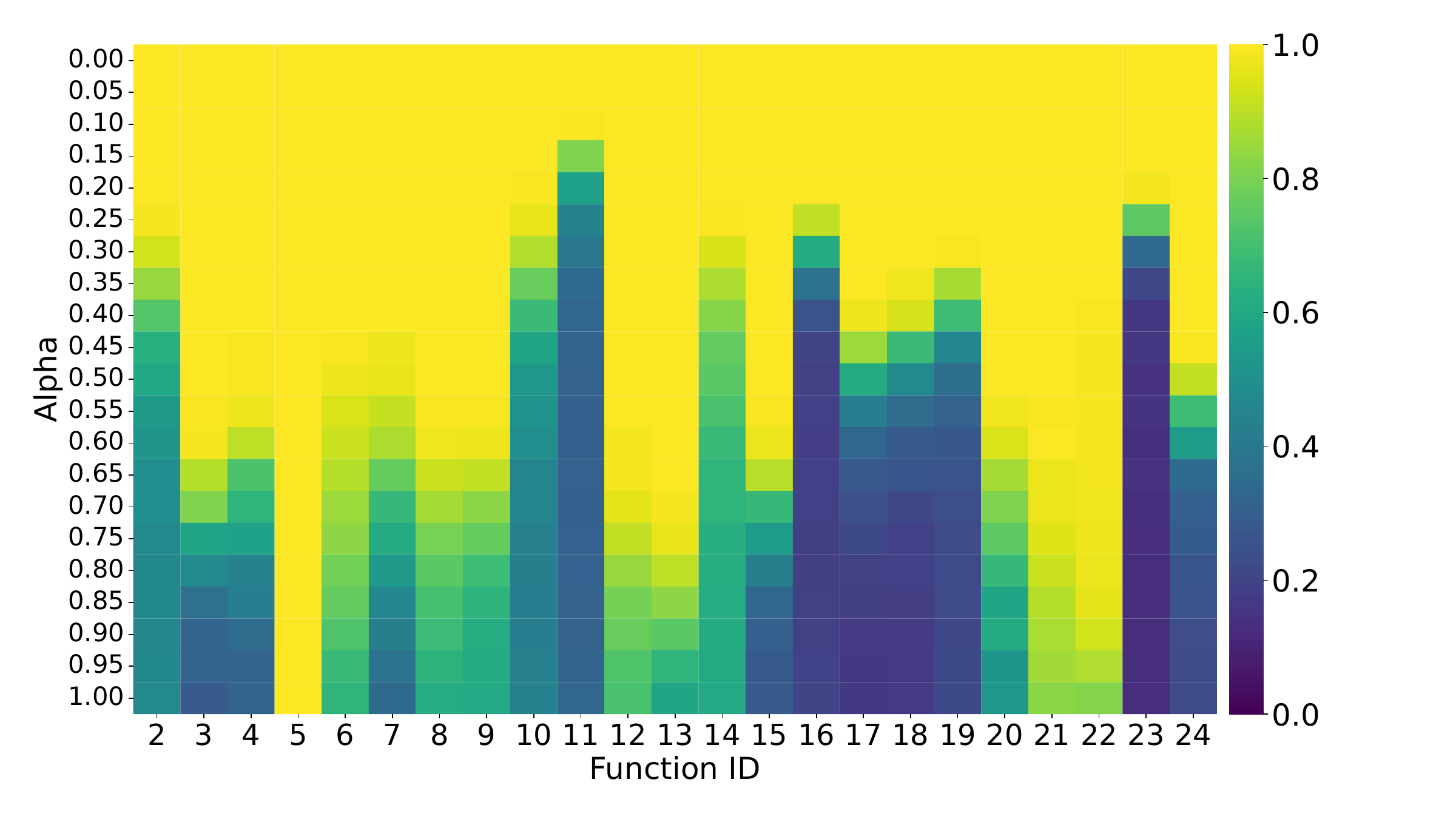}
        \caption{AOCC values for Cobyla.}  
        \label{fig:cobyla_heatmap_wsphere}
    \end{subfigure}
    \hfill
    \begin{subfigure}[b]{0.475\textwidth}   
        \centering 
        \includegraphics[width=\textwidth, trim=0mm 0mm 5mm 8mm,clip]{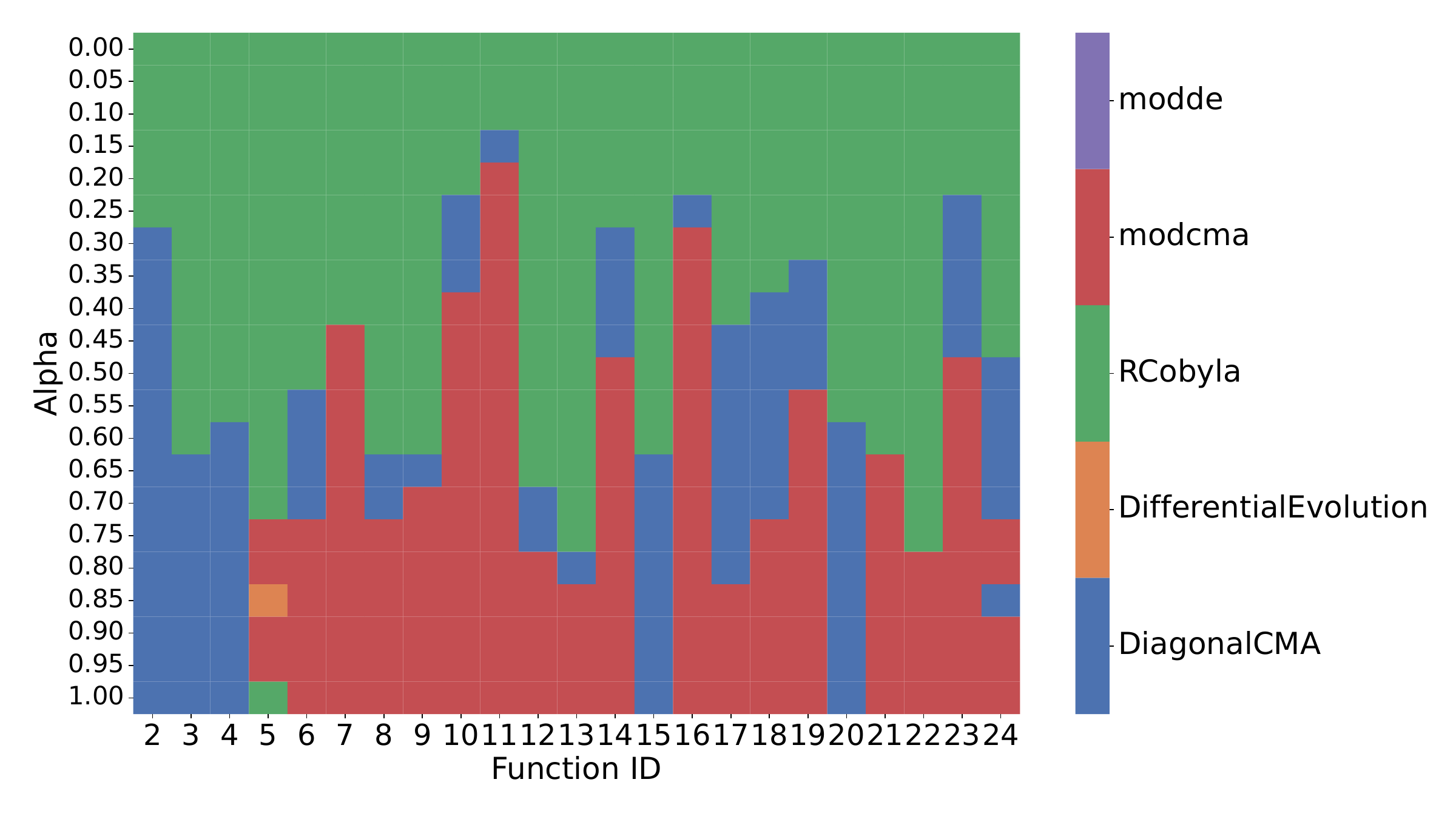}
        \caption{Best performing algorithm from the portfolio, based on AOCC.}    
        \label{fig:rank_grid}
    \end{subfigure}
    \caption{Normalized area under the ECDF curve of three selected algorithms (a-c) and best-ranking algorithm from the full portfolio (d) for each combination of the BBOB-function (x-axis) with a sphere model, for given values of $\alpha$ (y-axis). AOCC is calculated after $10\,000$ function evaluations, based on 50 runs on 50 instances (and location of optimum). Note that $\alpha=0$ corresponds to the sphere function.} 
    \label{fig:full_figure_aucs}
\end{figure*}

In Figure~\ref{fig:cma_heatmap_wsphere}, we can see that the performance of CMA-ES does indeed seem to move smoothly between the sphere and the function with which it is combined. It is however interesting to note the differences in speed at which this transition occurs. For example, while the final performance on functions 3 and 10 seems similar, the transition speed differs significantly. This seems to indicate that for F10, the addition of some global structure has a relatively weak influence on the challenges of this landscape from the perspective of the CMA-ES, while even small amounts of global structure significantly simplify the landscape of F3.

We can perform a similar analysis on other optimization algorithms. In Figures~\ref{fig:de_heatmap_wsphere} and~\ref{fig:cobyla_heatmap_wsphere}, we show the same heatmap as Figure~\ref{fig:cma_heatmap_wsphere}, but for Differential Evolution and Cobyla, respectively. It is clear from these heatmaps that the performance of DE is more variable than that of CMA-ES, while Cobyla's performance drops off much more quickly. The overall trendlines for DE do seem to be somewhat similar to those seen for diagonal CMA-ES: the transition points between high and low AOCC in Figure~\ref{fig:de_heatmap_wsphere} are comparable to those seen in Figure~\ref{fig:cma_heatmap_wsphere}. There are however still some differences in behavior, especially relative to Cobyla. These differences then lead to the question of whether there exist transition points in ranking between algorithms as well. Specifically, if one algorithm performs well for $\alpha=0$ but gets overtaken as $\alpha\rightarrow1$, exploring this change in ranking would give further insight into the relative strengths and weaknesses of the considered algorithms.  

To study the impact on the relative ranking of algorithms, we make use of the full portfolio of 5 algorithms and rank them based on AOCC on each affine function combination. We then visualize the top ranking algorithm on each setting in Figure~\ref{fig:rank_grid}. From this figure, we can see that Cobyla deals well with the sphere model, managing to outperform the other algorithms when the weighting of the sphere is relatively high. Then, after a certain threshold, the CMA-ES variants consistently outperform the rest of the portfolio, with dCMA taking over when Cobyla is no longer preferred. However, as $\alpha$ increases further, and the influence of the sphere model diminishes, an interesting pattern seems to occur. For several problems, there is a second transition point to modcma, indicating that the differences in default parameterizations between the used libraries have a large impact on the algorithms' behavior. One significant factor is related to the initial stepsize, which is smaller for dCMA, and thus might lead to it becoming more easily stuck in local optima when the global structure is not as strong.  

\begin{figure*}
    \centering
    \includegraphics[width=0.97\textwidth]{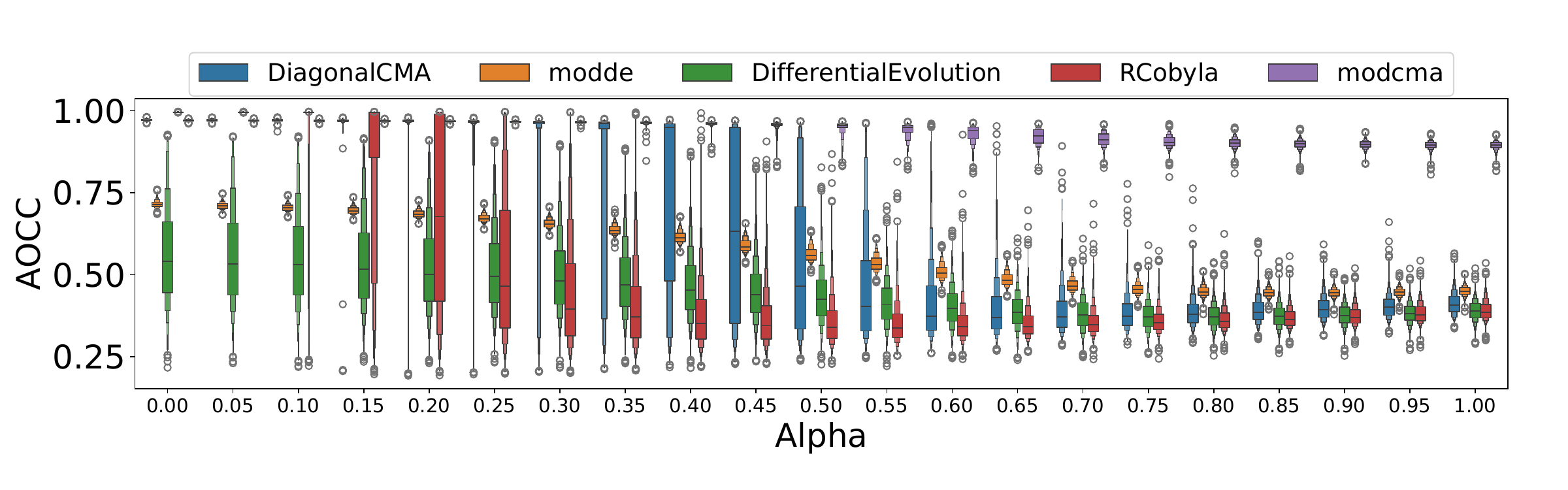}
    \caption{Distribution of AOCC values for 5 algorithms on the affine combinations between F10 
    ($\alpha =1 $) and F1 ($\alpha=0$), for selected values of $\alpha$.}
    \label{fig:auc_f10}
\end{figure*}

In order to better understand what the transitions in algorithm ranking look like, we can zoom in on one of the functions and plot the distribution of AOCC for all values of $\alpha$. This is done in Figure~\ref{fig:auc_f10}, where we look at the combination between F10 and the sphere model. In this figure, we observe that Cobyla is very effective at optimizing the sphere and the combinations with low $\alpha$. However, when $\alpha$ increases, Cobyla quickly starts to fail, while, for example, DiagonalCMA still manages to solve most instances at  $\alpha=0.25$ with similar AOCC. As $\alpha$ increases further, the modcma's performance remains stable, showing only a minor drop in performance relative to the one seen in dCMA. 

\subsection{Impact on ELA Features}

\begin{figure}
    \centering
    \includegraphics[width=\textwidth, trim=0mm 0mm 150mm 0mm,clip]{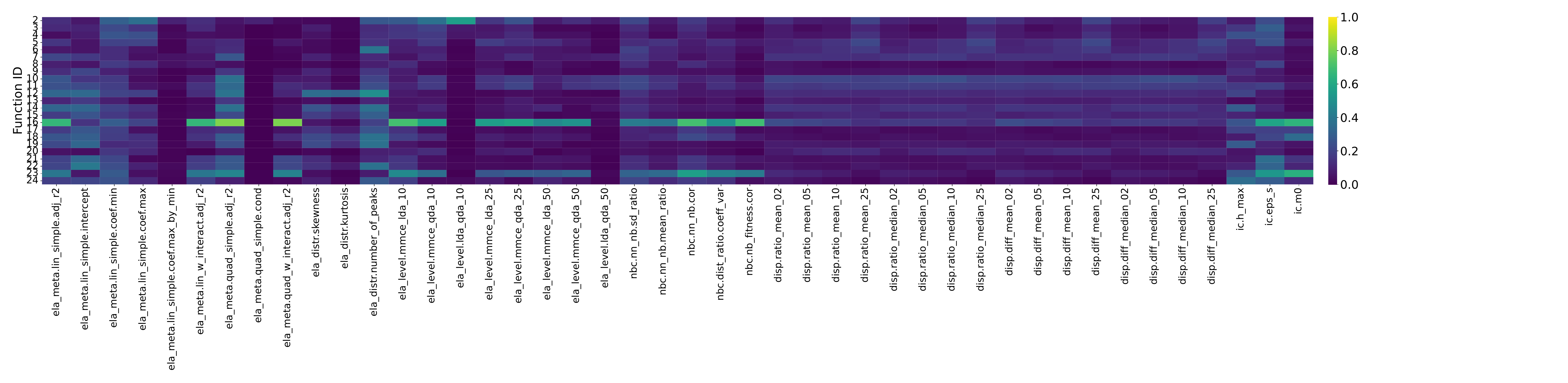}
    \caption{Total changes in each ELA feature when transitioning from the sphere to the function indicated in the row. Each cell represents the sum of differences in mean between pairs of consecutive values of $\alpha$, so high values indicate a large total change in mean, while low values indicate features which remain stable throughout the transition. }
    \label{fig:sphered_mean_diffs}
\end{figure}

In addition to the performance perspective, we can also look at what happens to the landscape feature of the BBOB functions as we add increasingly more influence from the sphere function. Since we measure $44$ different ELA features, our analysis of the impact is rather more high-level than the algorithm performance viewpoint, as we first aim to capture the overall stability of the features for increasing $\alpha$ values. This is measured as the sum of absolute differences in feature mean for consecutive $\alpha$'s, which is plotted in Figure~\ref{fig:sphered_mean_diffs}. From this figure, we can see that the mean of most features remains quite stable, with a few notable exceptions. In particular, functions 16 and 23 show many feature changes from the sphere, which matches observations from e.g.~\cite{Renau2019features, vskvorc2020understanding}. 

\begin{figure}
    \centering
    \includegraphics[width=\textwidth, trim=0mm 0mm 90mm 0mm,clip]{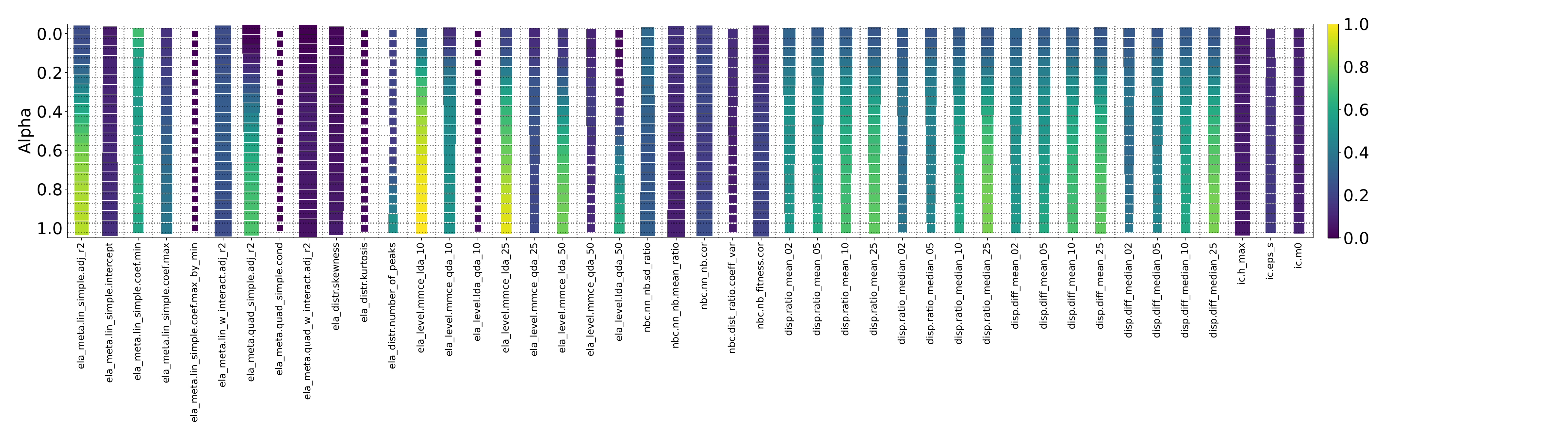}
    \caption{Evolution of ELA features with changing between Sphere ($\alpha = 0$) and F10 ($\alpha=1$). The color indicates the mean of the feature over the 50 instances (lighter = larger), while the size indicates the variance (larger = higher variance). }
    \label{fig:f10_mean_std_sphered}
\end{figure}

In addition to the differences between functions, it is also clear to see that features don't all behave in the same way. Since Figure~\ref{fig:sphered_mean_diffs} only shows absolute changes in mean, the deviation between instances might also play a role. To analyze this in some more detail, we select a single function (F10) and look at the evolution of both the feature mean and its standard deviation in Figure~\ref{fig:f10_mean_std_sphered}. From this figure, we can see that the mean of each feature seems to transition rather smoothly between the two component functions, in a similar way to the performance plots in Section~\ref{sec:sphere_perf}. However, we should note that the standard deviation of many features is relatively large for each $\alpha$ value. 

\subsection{Impact of Optimum Location and the Instance}\label{sec:loc_opt}

\begin{figure}
    \centering
    \includegraphics[width=0.4\textwidth]{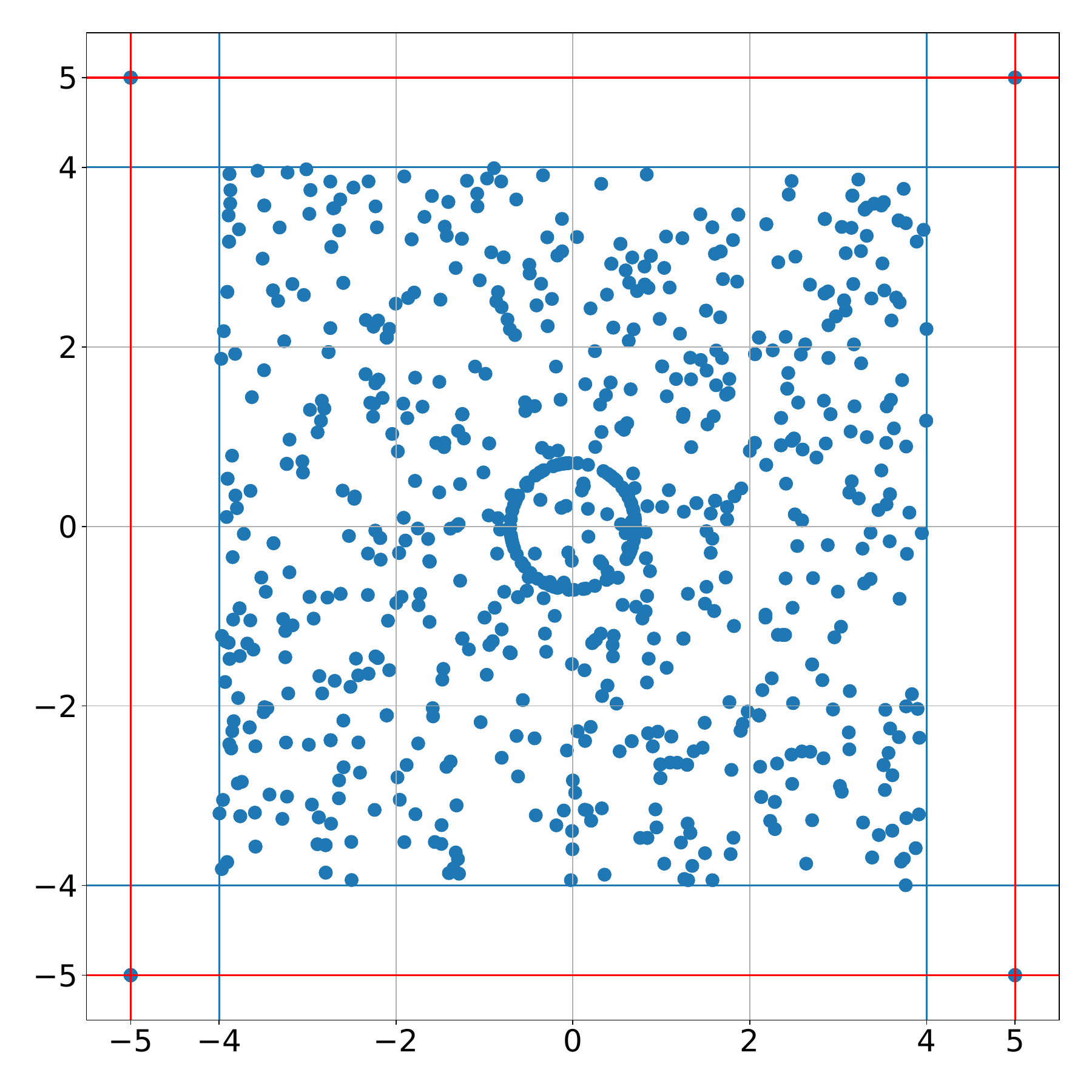}
    \caption{Location of optima of the 24 2d BBOB functions (1000 random instances). The red lines mark the commonly  used box-constraints of  $[-5,5]^d$. }
    \label{fig:2d_opt_loc}
\end{figure}

As can be seen from the relatively large variance in both ELA features and algorithm performance, the instance and location of the optimum can have a major impact on both the landscape and the corresponding algorithm behaviour. In particular, the way in which we defined an instance in our setup is not necessarily equivalent to the common interpretation, e.g., from BBOB. Since we allow the optimum of a component function to be moved anywhere in the domain, this can lead to large parts of the original function no longer being reachable. This is the main reason why some BBOB functions have very restricted distributions for their optima, which can be seen by analyzing the overall distribution of optima across all BBOB functions, visualized in Figure~\ref{fig:2d_opt_loc} and previously observed in e.g.\cite{evostar_bbob_instance}. 

\begin{figure}
    \centering
    \includegraphics[width=\textwidth]{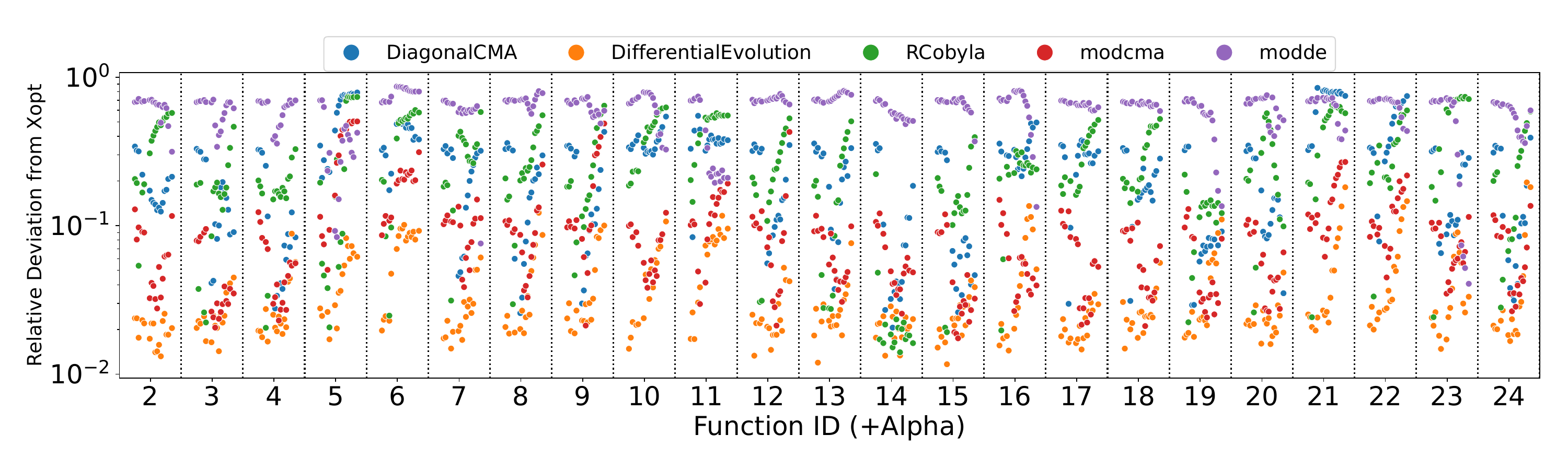}
    \caption{Relative deviation in AOCC caused by the change of the global optimum for all combinations of BBOB functions with the sphere model (calculated as deviation per instance divided by deviation across all instances). X-axis indicates changing function ID, and within each function ID the transition goes from $\alpha = 0$ to  $\alpha = 1$ (left to right).}
    \label{fig:instance_dev_rel}
\end{figure}

To analyse how much the algorithms in our portfolio are influenced by the choice of instance and location of optimum, we determine the relative impact of different instances for each function ($F_1$ and $\alpha$ value). This is done by first averaging the performance across all $50$ runs on each instance. By dividing this by the total variance present across all runs on all instances of that function, we obtain a relative measure of `stability' across instances, which is visualized in Figure~\ref{fig:instance_dev_rel}. This figure shows that some algorithms are inherently more impacted by the instance/location of the optimum (modde), while, for example, for Cobyla, the variance increases with increasing $\alpha$, which suggests that it is very stable on the sphere problem, but becomes much more impacted by variations in the landscape when more non-sphere influence is added. 

\begin{figure}
    \centering
    \includegraphics[width=\textwidth]{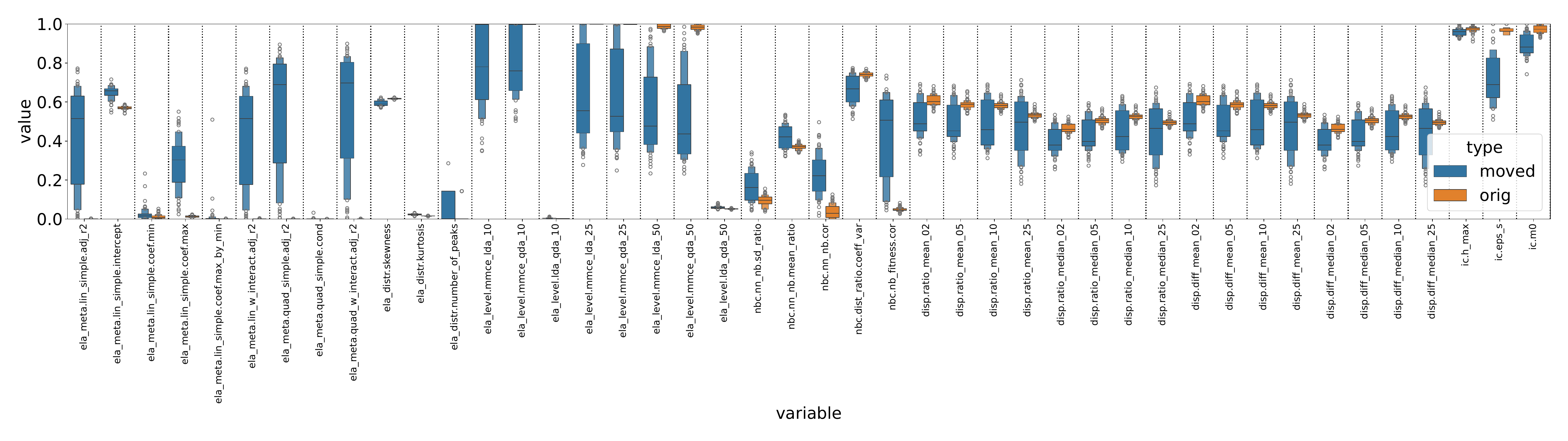}
    \caption{Distribution of normalized ELA features for the BBOB instance creation procedure and the same instance moved to have an optimum location uniformly in the domain, for F23. (other functions in the bbob vs moved folder)}
    \label{fig:bbob_distr_f23}
\end{figure}

To further analyse the impact of this increased flexibility in terms of function generation, we perform an experiment involving two versions of the original BBOB functions. The first is the data from Section~\ref{sec:sphere_perf} where $\alpha\in\{0,1\}$, which corresponds to data for 50 instances of each function, with each of them having its optimum moved to a random point in the domain. The second set of instances are created by taking the same instances of the BBOB functions, but not shifting their optimum (the rescaling from Section~\ref{sec:bbobscaling} is still applied). This allows us to compare the influence of moving the optimum on the landscape of the resulting problem. In Figure~\ref{fig:bbob_distr_f23}, we compare the distribution of all features on these two versions of BBOB function 23. 

\begin{figure}
    \centering
    \begin{subfigure}[][][t]{0.49\textwidth}
        \includegraphics[width=\textwidth]{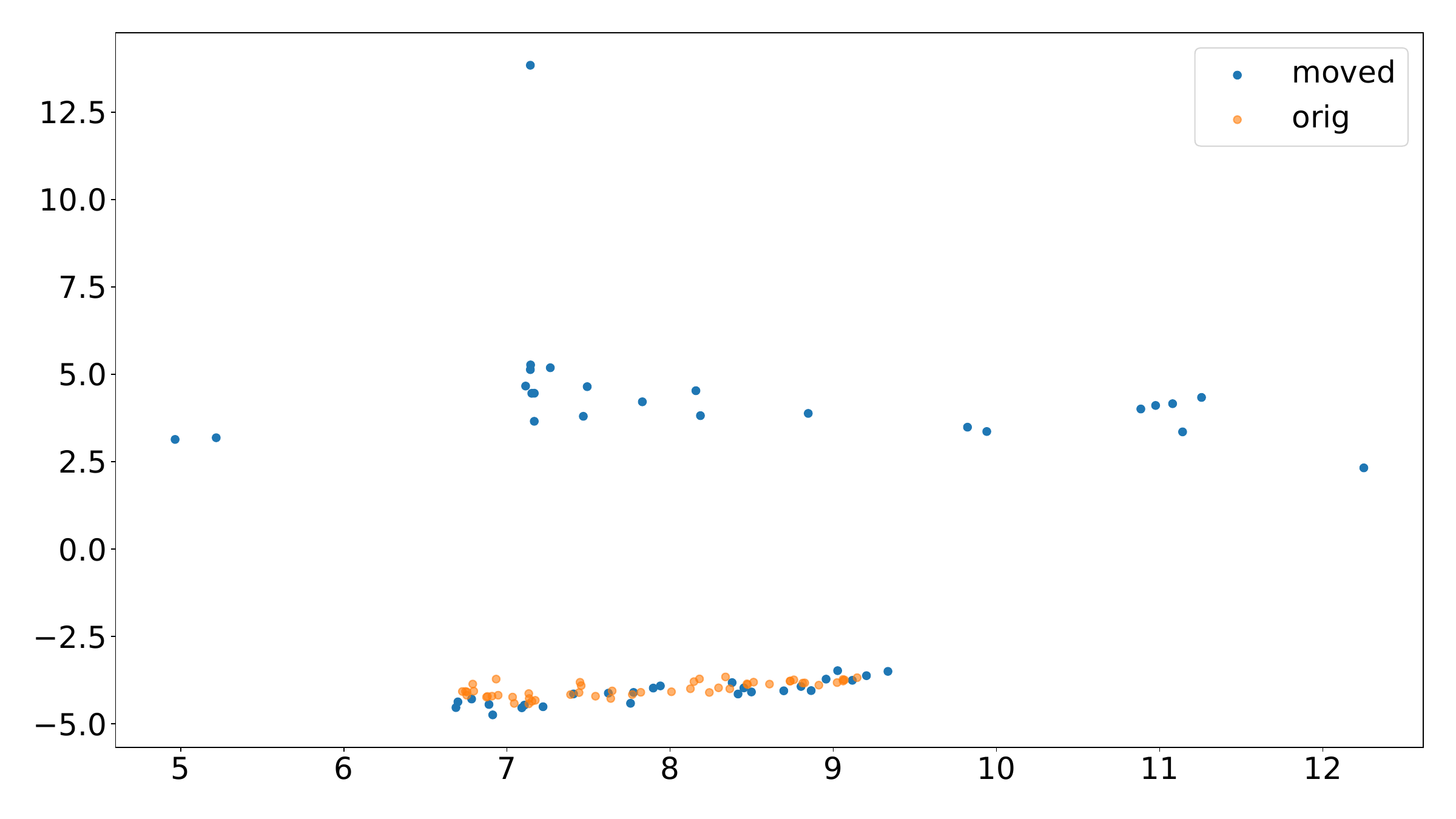}
    \caption{Projection of the moved and original version of F18 (50 instances).}
    \label{fig:umap_bbob_moved_f18}
    
    \end{subfigure}
    \begin{subfigure}[][][t]{0.49\textwidth}
            \includegraphics[width=\textwidth]{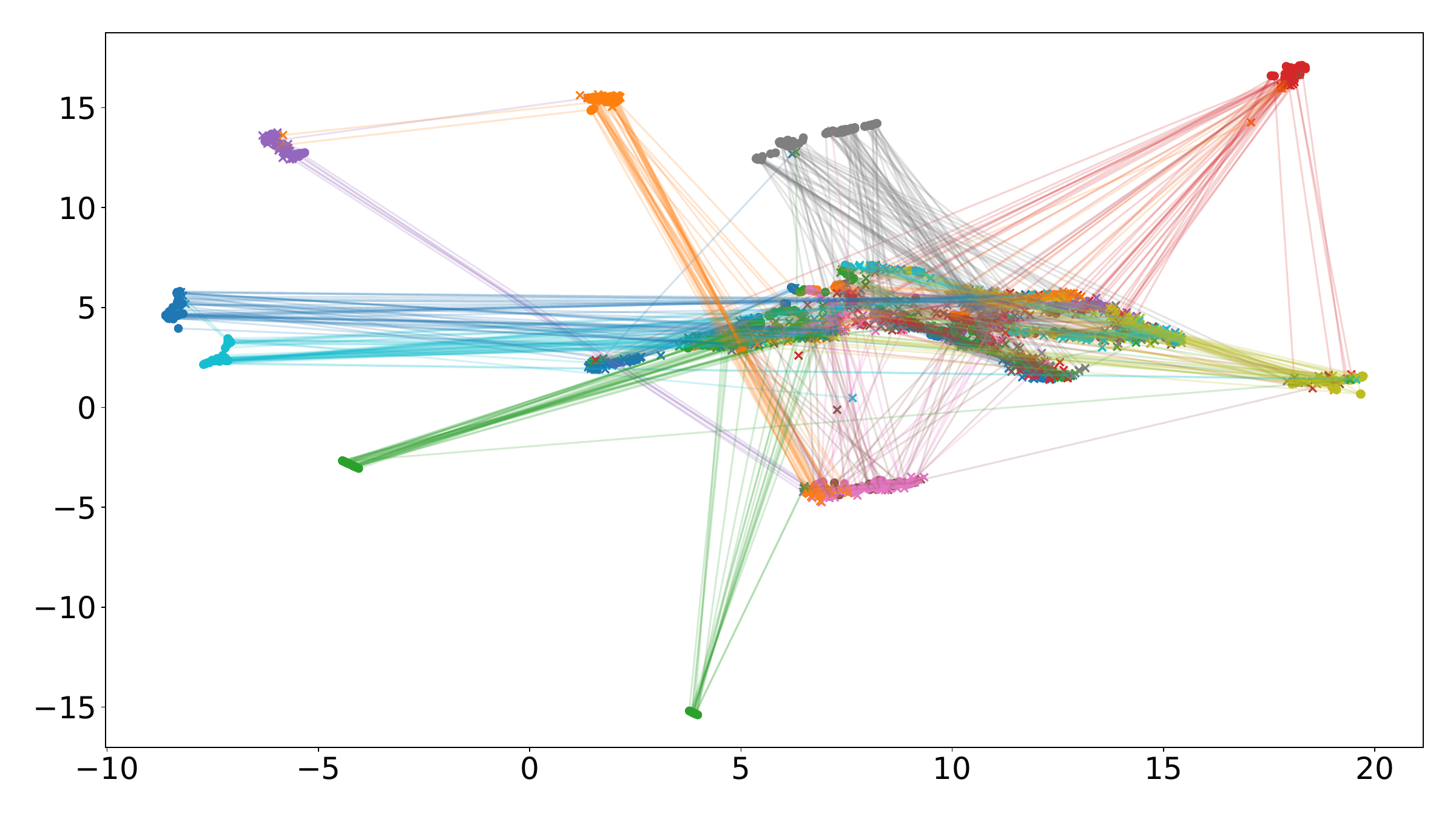}
    \caption{Projection of all BBOB functions connecting the original (circles) and moved (crosses) versions of each function.}
    \label{fig:umap_bbob_moved_all}
    \end{subfigure}
    \caption{UMAP projection trained on original bbob, then used to plot both the moved and original functions in 2D. }\label{fig:umap_moved}
\end{figure}

Finally, we can take an aggregated view of the features, and project the $44$ dimensional space into two dimensions using PCA on the original BBOB versions. Then, we can plot the moved versions of the function into the same space, and observe the differences. The result of this projection is shown in Figure~\ref{fig:umap_moved}, where we see that many functions are moved much closer to the center of the projected space. This suggests that some of the 'unique' feature combinations present in the original BBOB functions are being lost when moving their optimum. This happens because large parts of the function are moved outside of the domain, and replaced by parts which were originally located outside the bounds. For some functions, these components are exponentially increasing, leading to a large part of the space which is dominated by these artifacts, which is represented in the ELA-features. 

\subsection{Pairwise Combinations}\label{sec:pairs_funcgroups}

\begin{figure}
    \centering
    \includegraphics[width=\textwidth]{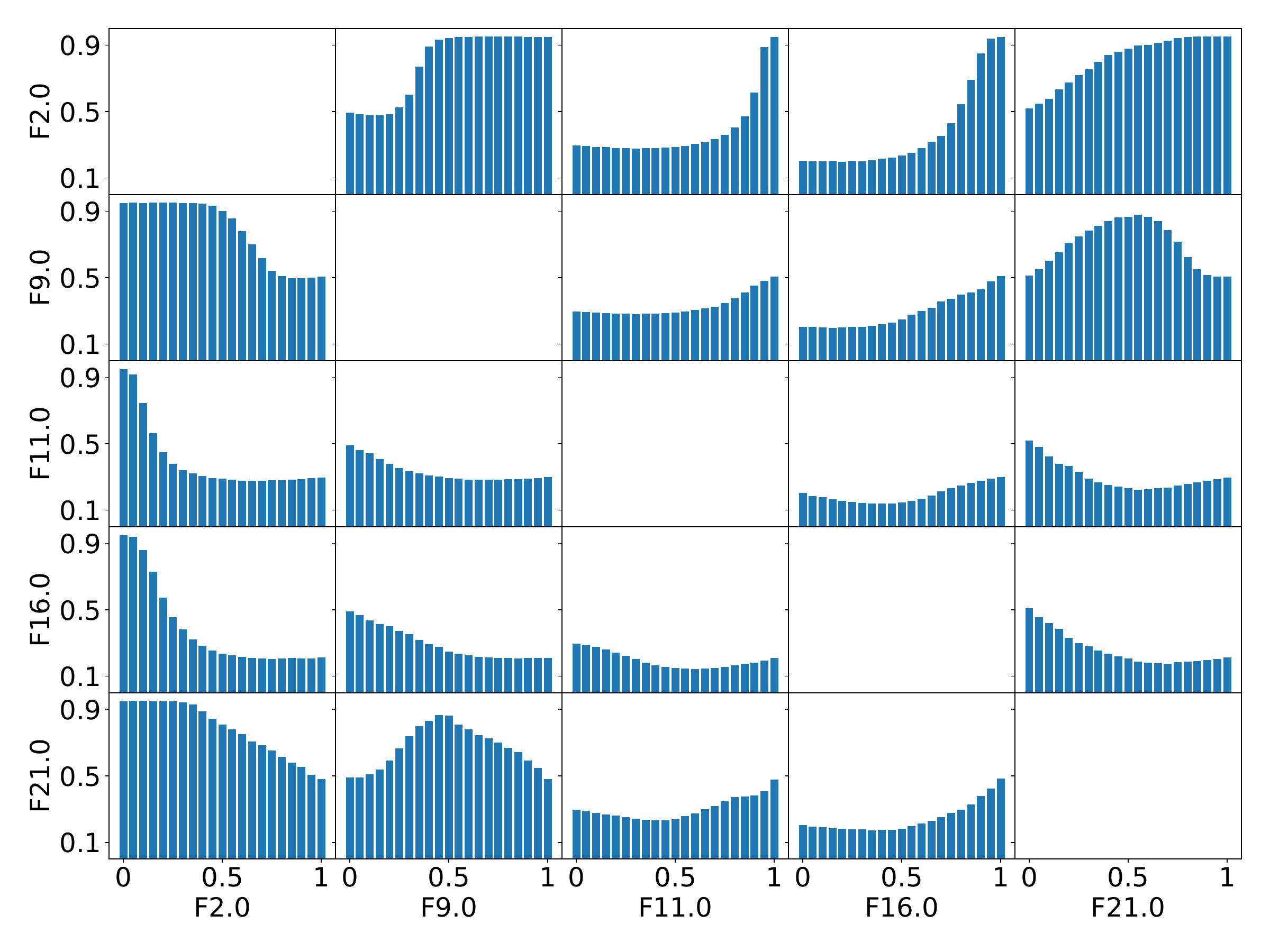}
    \caption{Normalized area over the convergence curve for Diagonal CMA-ES on each of the affine combinations between the selected BBOB problems. Each facet corresponds to the combination of the row and column function, with the x-axis indicating the used $\alpha$. AOCC values are calculated based on 50 runs on 25 instances, with a budget of $10\,000$ function evaluations.}
    \label{fig:cma_grid}
\end{figure}

While combining functions with a sphere model can be viewed as adding global structure to a problem, combinations between other functions can provide interesting insights into the transition points between different types of problems. To illustrate the kinds of insights that can be gained from these combinations, we select a subset of 5 functions and collect performance data on each combination with the same 21 $\alpha$ values (with both orderings of the function). We show the performance in terms of normalized AOCC of diagonal CMA-ES on these function combinations in Figure~\ref{fig:cma_grid}. Note that for $\alpha=1$, we are using the function specified in the column label, while for $\alpha=0$ we have the function specified in the row label.

When comparing this Figure to its equivalent from the GECCO paper~\cite{ABBOB-GECCO}, it is important to note the fact that Figure~\ref{fig:cma_grid} is almost fully symmetric around the diagonal, which was not the case in the GECCO paper. Even though we might expect $(F_1, F_2, \alpha)$ to be similar to $(F_2, F_1, 1-\alpha)$, this was not the case when the location of the global optimum was selected as the optimum of one of the component functions, as different BBOB functions can have significantly different distributions of potential global optima~\cite{evostar_bbob_instance}. This is in large part the reason why we enabled the MA-BBOB generator to sample the optimum uniformly at random in the domain. While Section~\ref{sec:loc_opt} showed that this can potentially move interesting parts of some component functions outside the domain, we view this as a worthwhile tradeoff to achieve fully unbiased global optima. 

From Figure~\ref{fig:cma_grid}, we can also see that the transition of performance between the two extreme $\alpha$ values is mostly smooth. While there are some rather quick changes, e.g., for the transition between F2 and F11, these seem to be the exception rather than the rule. Particularly interesting are the settings where the performance of affine combinations between two functions proves to be much easier or harder than the functions which are being combined. For example, this is the case for the combinations of F21 and F9. 

\section{Combining Multiple Functions: Testing Generalizability}\label{sec:alg_sel}

For our final set of experiments, we make use of a set of $1000$ functions generated using the setup described in Section~\ref{sub:weights}. This data is taken directly from the Zenodo repository~\cite{zenodo_automl}, and contains both ELA and performance data (for the same set of algorithms described in Section~\ref{sec:setup1}, but using the original AUC measure instead of the AOCC). In~\cite{automl_mabbob} we analyzed this data to understand the MA-BBOB instance generation procedure, with the goal of generating a wide set of benchmark problems on which algorithm selection and other automated machine learning techniques can be tested.  

\begin{figure}
    \centering
    \begin{subfigure}[][][t]{0.49\textwidth}
        \includegraphics[width=\textwidth]{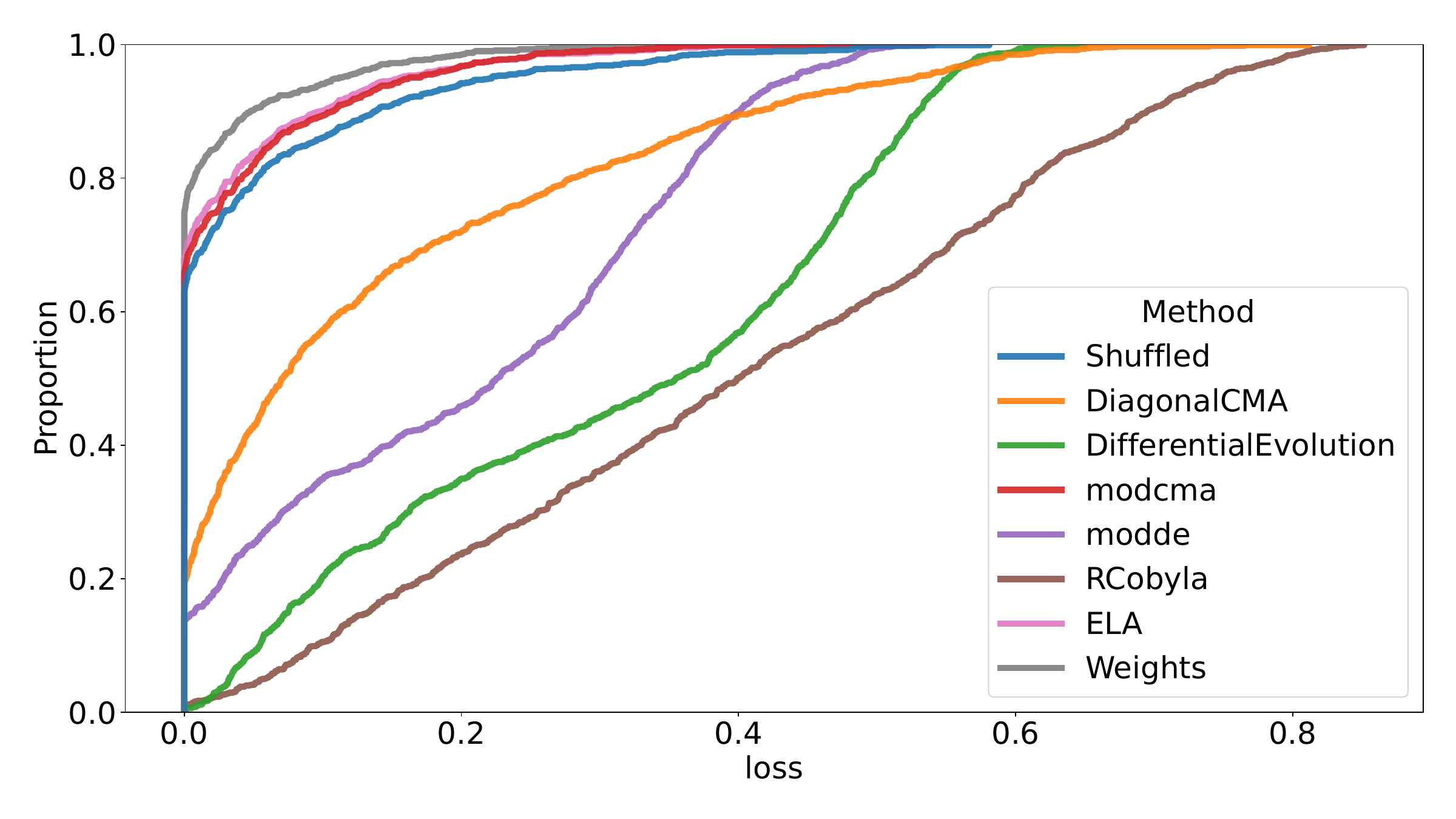}
    \caption{AUC loss for 5 dimensional functions.}
    \label{fig:ecdf_5d}
    \end{subfigure}
    \begin{subfigure}[][][t]{0.49\textwidth}
            \includegraphics[width=\textwidth]{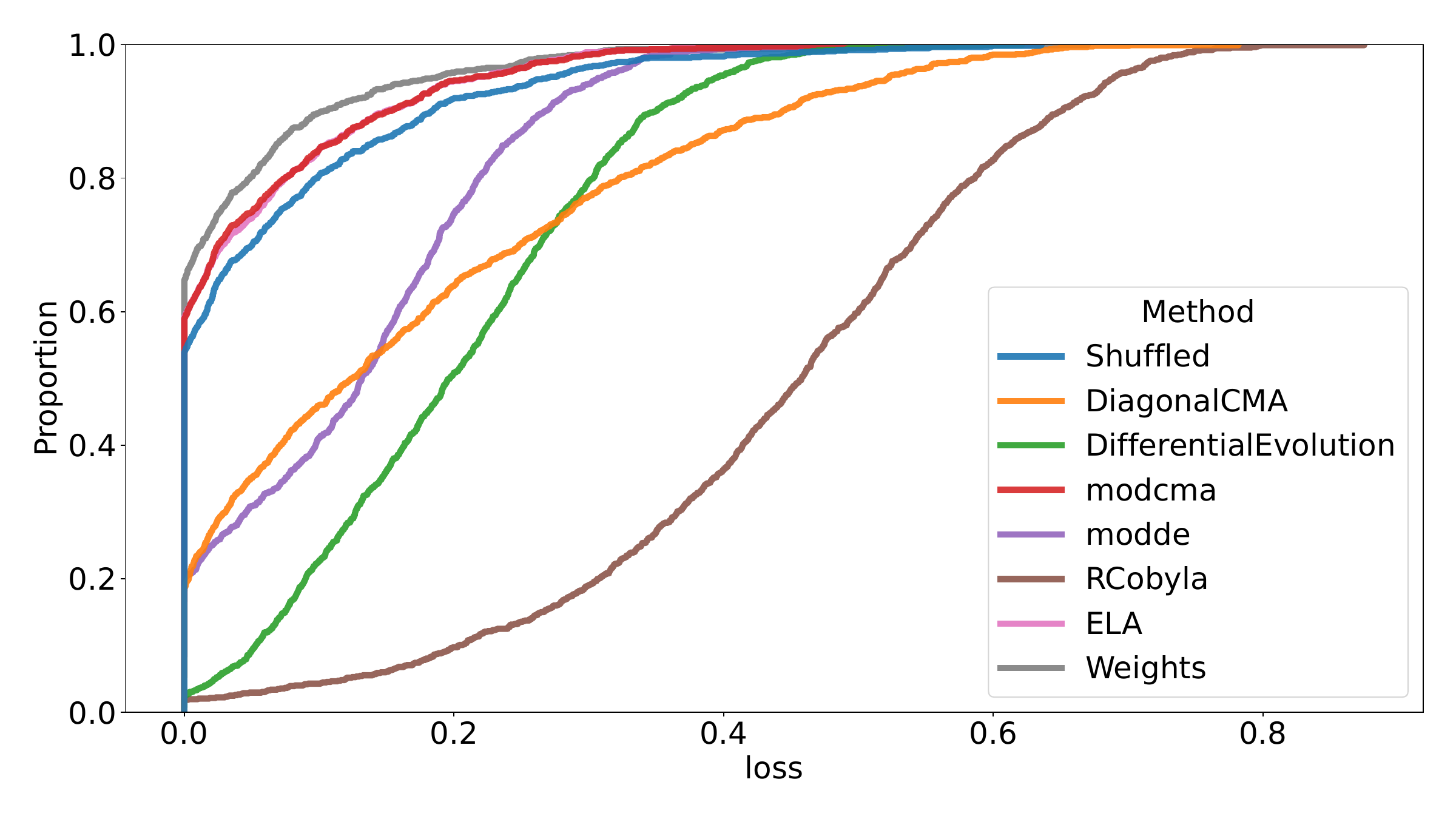}
    \caption{AUC loss for 2 dimensional functions.}
    \label{fig:ecdf_2d}
    \end{subfigure}
    \caption{Cumulative loss (AUC) for different models: cross-validation (mixture of BBOB + MA-BBOB generated combinations) based on weights and ELA, and each of the single-algorithm models.}\label{ecdf1}
\end{figure}

In this experiment, we take the perspective of algorithm selection and train a random forest model to predict the best algorithm to use for each function, based either on the ELA features of the problem or the weights of the component functions. We can then compare the loss in terms of AUC relative to the virtual best solver (VBS) for both of these models, in different training contexts. We can either use the common cross-validation setup, or attempt to test for generalization ability based only on the original BBOB functions. In Figure~\ref{fig:ecdf_5d} we show the cumulative loss for the cross-validation setup on the 5-dimensional functions. From this, we can see that the ELA-based selector performs worse than the one based on the weights. This confirms the previous observation that the ELA features might not be sufficiently representative to accurately represent the problems in a way which is relevant for ranking optimization algorithms. 

In order to better estimate how much the structure of the ELA features helps the prediction, we can add in a naive baseline. This is created by shuffling the labels (best ranked algorithm) of all samples before training. This shuffled model is in essence just a selector based on the frequency of labels in the training data, and the difference between this version and the original ELA-based selector shows how much the structure of the ELA-features helps improve the predictions. The results for the cross-validation setup in 2D are shown in Figure~\ref{fig:ecdf_2d}, where we see that the benefit over most of the individual algorithms is inherent to the selected portfolio, since the shuffled model outperforms all algorithms except modcma. This suggests that our algorithm portfolio is severely unbalanced.  

\begin{figure}
    \centering
    \begin{minipage}{.45\textwidth}
    \includegraphics[width=\textwidth]{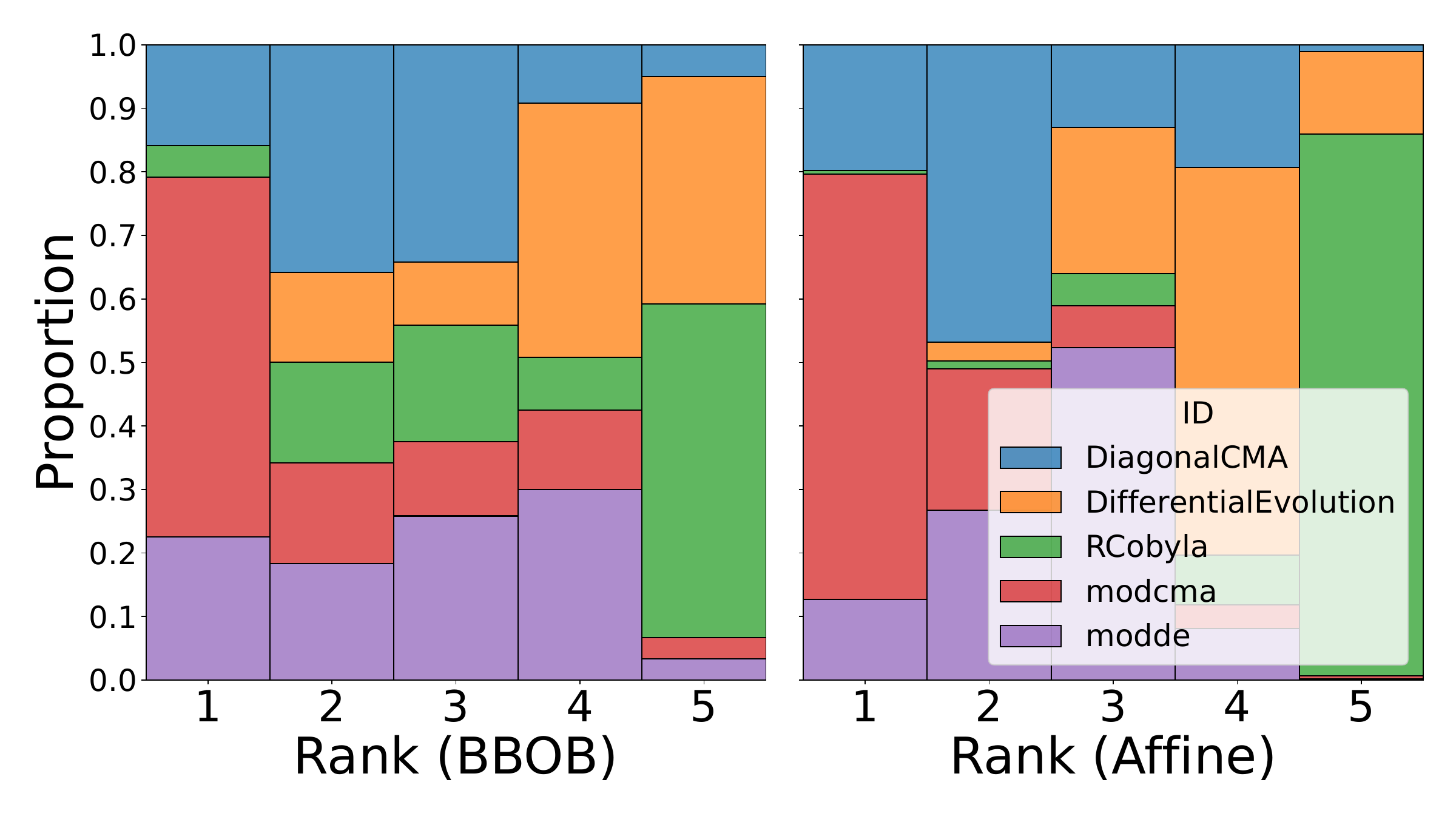}
    \caption{Distribution of ranks based on per-function AUC after $10\,000$ evaluations. }
    \label{fig:auc_ranking}
    \end{minipage}\hspace{0.05\textwidth}
    \begin{minipage}{.45\textwidth}
    \includegraphics[width=\textwidth]{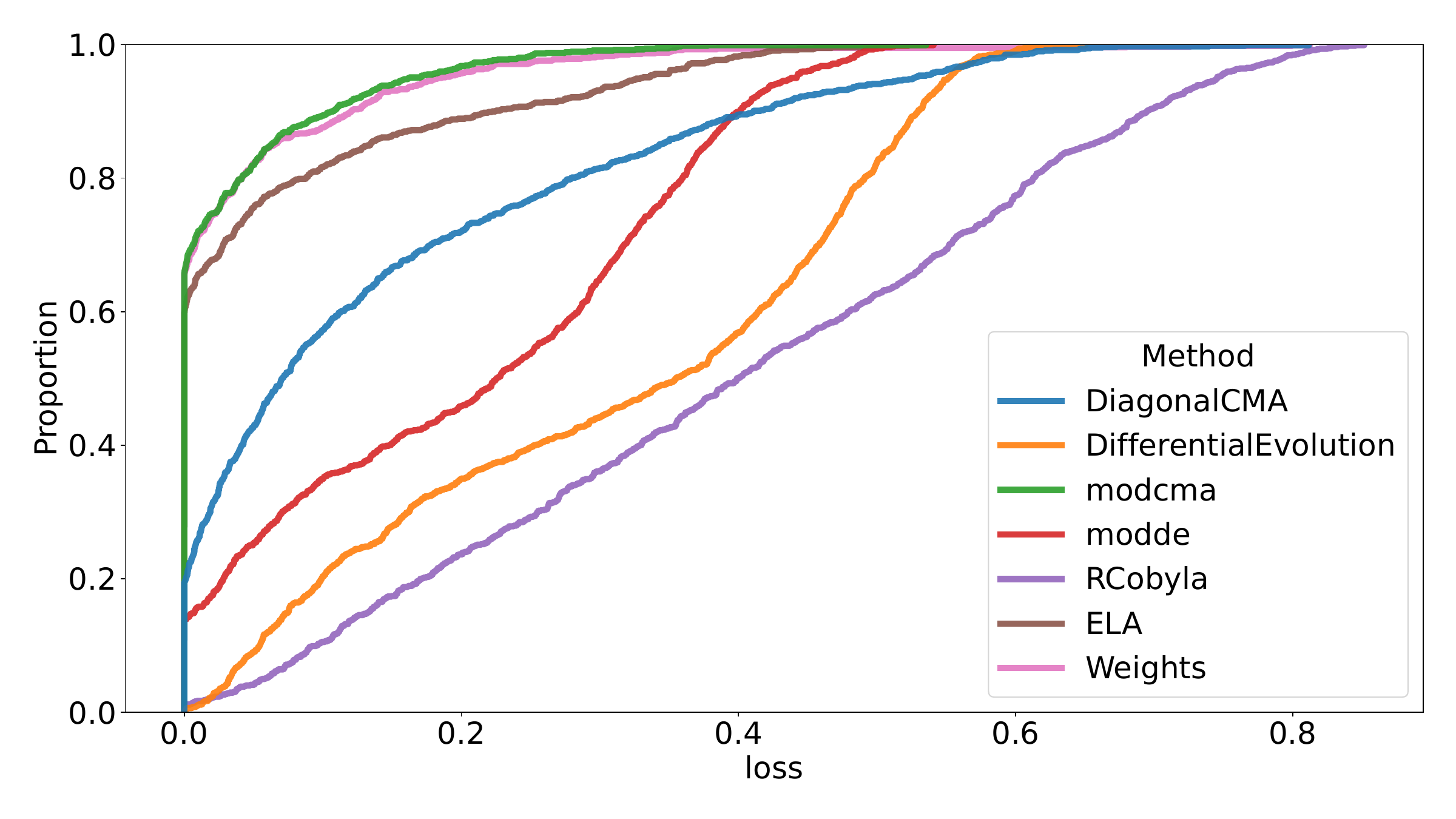}
    \caption{Cumulative loss (AUC) on the 5 dimensional MA-BBOB problems for models trained on the BBOB functions, and each of the single-algorithm models.}
    \label{fig:ecdf_generalize_5d}
    \end{minipage}
\end{figure}

When looking at the generalization task, this imbalance is exacerbated further, since for MA-BBOB the modcma is ranked first on an even larger fraction of functions than on BBOB, as shown in Figure~\ref{fig:auc_ranking}. In combination with the added challenge of transferring to a new suite, this leads to our algorithm selection models being outperformed by the modcma, which is the Single Best Solver (SBS) in this case, as illustrated in Figure~\ref{fig:ecdf_generalize_5d}. While the algorithm portfolio is partly responsible for this shortcoming, the generalization ability does not significantly improve when removing the modcma from our portfolio. This suggests that training on the original BBOB instances does not sufficiently represent the challenges faced in the MA-BBOB suite. An important aspect of the challenge of this transfer is the location of the optima, as discussed in Section~\ref{sec:loc_opt}.

\section{Conclusions and Future Work}

We propose MA-BBOB as a way to generate new benchmark problems by creating affine combinations of rescaled and translated BBOB functions. Our approach generalizes  earlier work by Dietrich and Mersmann who investigated combinations of pairs of BBOB functions and the transitions between them~\cite{affinebbob}. The construction of MA-BBOB still allows for detailed studies into the relation between landscape features and algorithm performance, but at the same time, it has the potential to be used for testing algorithm selection techniques for optimization algorithms. 

While MA-BBOB does not bias the location of the optimum to any part of the domain, this comes at a potential cost in terms of instance creation. Specifically, we have shown that using this procedure, the landscape features of the original BBOB features can be drastically changed. Because of this, we shouldn't consider two MA-BBOB functions as instances of the same problem if they share the same weights, but view them as different problems instead. 

When analyzing the changes in both performance and landscape of combinations of two functions, we observe relatively smooth transitions between the components. There is however a large variance in both of these metrics. It is additionally worthwhile to note that combinations between two functions might become easier or harder for a given optimization algorithm to solve. Further studies into these transitions might help reveal some closer relation between the performance on the components and on their affine combination, and link this with the changes in landscape characteristics. 

The selection of `representative' instance collections still remains to be done. Another important step for future work is to test the generalization ability of AutoML systems that are trained on MA-BBOB functions and tested on numerical black-box optimization problems that do not originate from the BBOB family. In this context, our basic Random Forest-based algorithm selector indicates that the ELA features might not be as suitable for this generalization task as expected, motivating further research on feature engineering for single-objective continuous black-box optimization.

\begin{acks}
    This work was realized with the support of the ANR T-ERC project \emph{VARIATION} (ANR-22-ERCS-0003-01) and by the CNRS INS2I project \emph{IOHprofiler}.
\end{acks}

\bibliographystyle{ACM-Reference-Format}
\bibliography{sample-base}


\begin{thebibliography}{40}


\ifx \showCODEN    \undefined \def \showCODEN     #1{\unskip}     \fi
\ifx \showDOI      \undefined \def \showDOI       #1{#1}\fi
\ifx \showISBNx    \undefined \def \showISBNx     #1{\unskip}     \fi
\ifx \showISBNxiii \undefined \def \showISBNxiii  #1{\unskip}     \fi
\ifx \showISSN     \undefined \def \showISSN      #1{\unskip}     \fi
\ifx \showLCCN     \undefined \def \showLCCN      #1{\unskip}     \fi
\ifx \shownote     \undefined \def \shownote      #1{#1}          \fi
\ifx \showarticletitle \undefined \def \showarticletitle #1{#1}   \fi
\ifx \showURL      \undefined \def \showURL       {\relax}        \fi
\providecommand\bibfield[2]{#2}
\providecommand\bibinfo[2]{#2}
\providecommand\natexlab[1]{#1}
\providecommand\showeprint[2][]{arXiv:#2}

\bibitem[Auger and Hansen(2020)]%
        {10yearsBBOB}
\bibfield{author}{\bibinfo{person}{Anne Auger} {and} \bibinfo{person}{Nikolaus
  Hansen}.} \bibinfo{year}{2020}\natexlab{}.
\newblock \bibinfo{title}{A SIGEVO Impact Award for a Paper Arising from the
  COCO Platform: A Summary and Beyond}.
\newblock
  \bibinfo{howpublished}{\url{https://evolution.sigevo.org/issues/HTML/sigevolution-13-4/home.html}}.
\newblock
Issue 3.


\bibitem[Bischl et~al\mbox{.}(2012)]%
        {bischl2012algorithm}
\bibfield{author}{\bibinfo{person}{Bernd Bischl}, \bibinfo{person}{Olaf
  Mersmann}, \bibinfo{person}{Heike Trautmann}, {and} \bibinfo{person}{Mike
  Preu{\ss}}.} \bibinfo{year}{2012}\natexlab{}.
\newblock \showarticletitle{Algorithm selection based on exploratory landscape
  analysis and cost-sensitive learning}. In
  \bibinfo{booktitle}{\emph{Proceedings of the 14th annual conference on
  Genetic and evolutionary computation}}. \bibinfo{pages}{313--320}.
\newblock


\bibitem[Cenikj et~al\mbox{.}(2022)]%
        {cenikj2022selector}
\bibfield{author}{\bibinfo{person}{Gjorgjina Cenikj},
  \bibinfo{person}{Ryan~Dieter Lang}, \bibinfo{person}{Andries~Petrus
  Engelbrecht}, \bibinfo{person}{Carola Doerr}, \bibinfo{person}{Peter
  Koro{\v{s}}ec}, {and} \bibinfo{person}{Tome Eftimov}.}
  \bibinfo{year}{2022}\natexlab{}.
\newblock \showarticletitle{{SELECTOR:} selecting a representative benchmark
  suite for reproducible statistical comparison}. In
  \bibinfo{booktitle}{\emph{Proc. of Genetic and Evolutionary Computation
  Conference (GECCO)}}. \bibinfo{pages}{620--629}.
\newblock


\bibitem[de~Nobel et~al\mbox{.}(2021a)]%
        {nobel_modcma_assessing}
\bibfield{author}{\bibinfo{person}{Jacob de Nobel}, \bibinfo{person}{Diederick
  Vermetten}, \bibinfo{person}{Hao Wang}, \bibinfo{person}{Carola Doerr}, {and}
  \bibinfo{person}{Thomas B{\"{a}}ck}.} \bibinfo{year}{2021}\natexlab{a}.
\newblock \showarticletitle{Tuning as a means of assessing the benefits of new
  ideas in interplay with existing algorithmic modules}. In
  \bibinfo{booktitle}{\emph{Proc. of Genetic and Evolutionary Computation
  Conference (GECCO'21, Companion material)}}. \bibinfo{publisher}{{ACM}},
  \bibinfo{pages}{1375--1384}.
\newblock
\urldef\tempurl%
\url{https://doi.org/10.1145/3449726.3463167}
\showDOI{\tempurl}


\bibitem[de~Nobel et~al\mbox{.}(2021b)]%
        {iohexp}
\bibfield{author}{\bibinfo{person}{Jacob de Nobel}, \bibinfo{person}{Furong
  Ye}, \bibinfo{person}{Diederick Vermetten}, \bibinfo{person}{Hao Wang},
  \bibinfo{person}{Carola Doerr}, {and} \bibinfo{person}{Thomas B{\"{a}}ck}.}
  \bibinfo{year}{2021}\natexlab{b}.
\newblock \showarticletitle{IOHexperimenter: Benchmarking Platform for
  Iterative Optimization Heuristics}.
\newblock \bibinfo{journal}{\emph{CoRR}}  \bibinfo{volume}{abs/2111.04077}
  (\bibinfo{year}{2021}).
\newblock
\showeprint[arXiv]{2111.04077}
\urldef\tempurl%
\url{https://arxiv.org/abs/2111.04077}
\showURL{%
\tempurl}


\bibitem[Dietrich and Mersmann(2022)]%
        {affinebbob}
\bibfield{author}{\bibinfo{person}{Konstantin Dietrich} {and}
  \bibinfo{person}{Olaf Mersmann}.} \bibinfo{year}{2022}\natexlab{}.
\newblock \showarticletitle{Increasing the Diversity of Benchmark Function Sets
  Through Affine Recombination}. In \bibinfo{booktitle}{\emph{Proc. of Parallel
  Problem Solving from Nature (PPSN'22)}} \emph{(\bibinfo{series}{LNCS},
  Vol.~\bibinfo{volume}{13398})},
  \bibfield{editor}{\bibinfo{person}{G{\"{u}}nter Rudolph},
  \bibinfo{person}{Anna~V. Kononova}, \bibinfo{person}{Hern{\'{a}}n~E.
  Aguirre}, \bibinfo{person}{Pascal Kerschke}, \bibinfo{person}{Gabriela
  Ochoa}, {and} \bibinfo{person}{Tea Tusar}} (Eds.).
  \bibinfo{publisher}{Springer}, \bibinfo{pages}{590--602}.
\newblock
\urldef\tempurl%
\url{https://doi.org/10.1007/978-3-031-14714-2\_41}
\showDOI{\tempurl}


\bibitem[Doerr et~al\mbox{.}(2018)]%
        {IOHprofilerArxiv}
\bibfield{author}{\bibinfo{person}{Carola Doerr}, \bibinfo{person}{Hao Wang},
  \bibinfo{person}{Furong Ye}, \bibinfo{person}{Sander van Rijn}, {and}
  \bibinfo{person}{Thomas B{\"{a}}ck}.} \bibinfo{year}{2018}\natexlab{}.
\newblock \showarticletitle{IOHprofiler: {A} Benchmarking and Profiling Tool
  for Iterative Optimization Heuristics}.
\newblock \bibinfo{journal}{\emph{CoRR}}  \bibinfo{volume}{abs/1810.05281}
  (\bibinfo{year}{2018}).
\newblock
\showeprint[arXiv]{1810.05281}
\urldef\tempurl%
\url{http://arxiv.org/abs/1810.05281}
\showURL{%
\tempurl}


\bibitem[Hansen et~al\mbox{.}(2021)]%
        {hansen2020coco}
\bibfield{author}{\bibinfo{person}{Nikolaus Hansen}, \bibinfo{person}{Anne
  Auger}, \bibinfo{person}{Raymond Ros}, \bibinfo{person}{Olaf Mersmann},
  \bibinfo{person}{Tea Tu{\v{s}}ar}, {and} \bibinfo{person}{Dimo Brockhoff}.}
  \bibinfo{year}{2021}\natexlab{}.
\newblock \showarticletitle{{COCO:} A platform for comparing continuous
  optimizers in a black-box setting}.
\newblock \bibinfo{journal}{\emph{Optim. Methods Softw.}} \bibinfo{volume}{36},
  \bibinfo{number}{1} (\bibinfo{year}{2021}), \bibinfo{pages}{114--144}.
\newblock


\bibitem[Hansen et~al\mbox{.}(2009)]%
        {bbobfunctions}
\bibfield{author}{\bibinfo{person}{Nikolaus Hansen}, \bibinfo{person}{Steffen
  Finck}, \bibinfo{person}{Raymond Ros}, {and} \bibinfo{person}{Anne Auger}.}
  \bibinfo{year}{2009}\natexlab{}.
\newblock \bibinfo{booktitle}{\emph{{Real-Parameter Black-Box Optimization
  Benchmarking 2009: Noiseless Functions Definitions}}}.
\newblock \bibinfo{type}{{T}echnical {R}eport} RR-6829.
  \bibinfo{institution}{{INRIA}}.
\newblock
\urldef\tempurl%
\url{https://hal.inria.fr/inria-00362633/document}
\showURL{%
\tempurl}


\bibitem[Hansen and Ostermeier(2001)]%
        {hansen2001self_adaptation_es}
\bibfield{author}{\bibinfo{person}{Nikolaus Hansen} {and}
  \bibinfo{person}{Andreas Ostermeier}.} \bibinfo{year}{2001}\natexlab{}.
\newblock \showarticletitle{Completely Derandomized Self-Adaptation in
  Evolution Strategies}.
\newblock \bibinfo{journal}{\emph{Evolutionary Computation}}
  \bibinfo{volume}{9}, \bibinfo{number}{2} (\bibinfo{year}{2001}),
  \bibinfo{pages}{159--195}.
\newblock
\urldef\tempurl%
\url{https://doi.org/10.1162/106365601750190398}
\showDOI{\tempurl}


\bibitem[Kerschke et~al\mbox{.}(2019)]%
        {KerschkeHNT19survey}
\bibfield{author}{\bibinfo{person}{Pascal Kerschke}, \bibinfo{person}{Holger~H.
  Hoos}, \bibinfo{person}{Frank Neumann}, {and} \bibinfo{person}{Heike
  Trautmann}.} \bibinfo{year}{2019}\natexlab{}.
\newblock \showarticletitle{Automated Algorithm Selection: Survey and
  Perspectives}.
\newblock \bibinfo{journal}{\emph{Evol. Comput.}} \bibinfo{volume}{27},
  \bibinfo{number}{1} (\bibinfo{year}{2019}), \bibinfo{pages}{3--45}.
\newblock
\urldef\tempurl%
\url{https://doi.org/10.1162/evco\_a\_00242}
\showDOI{\tempurl}


\bibitem[Kerschke and Trautmann(2019)]%
        {KerschkeT19}
\bibfield{author}{\bibinfo{person}{P. Kerschke} {and} \bibinfo{person}{H.
  Trautmann}.} \bibinfo{year}{2019}\natexlab{}.
\newblock \showarticletitle{Automated Algorithm Selection on Continuous
  Black-Box Problems by Combining Exploratory Landscape Analysis and Machine
  Learning}.
\newblock \bibinfo{journal}{\emph{Evolutionary Computation}}
  \bibinfo{volume}{27}, \bibinfo{number}{1} (\bibinfo{year}{2019}),
  \bibinfo{pages}{99--127}.
\newblock
\urldef\tempurl%
\url{https://doi.org/10.1162/evco\_a\_00236}
\showDOI{\tempurl}


\bibitem[Kostovska et~al\mbox{.}(2022)]%
        {AnjaPPSN2022}
\bibfield{author}{\bibinfo{person}{Ana Kostovska}, \bibinfo{person}{Anja
  Jankovic}, \bibinfo{person}{Diederick Vermetten}, \bibinfo{person}{Jacob de
  Nobel}, \bibinfo{person}{Hao Wang}, \bibinfo{person}{Tome Eftimov}, {and}
  \bibinfo{person}{Carola Doerr}.} \bibinfo{year}{2022}\natexlab{}.
\newblock \showarticletitle{Per-run Algorithm Selection with Warm-starting
  using Trajectory-based Features}. In \bibinfo{booktitle}{\emph{Proc. of
  Parallel Problem Solving from Nature (PPSN'22)}}
  \emph{(\bibinfo{series}{LNCS}, Vol.~\bibinfo{volume}{13398})}.
  \bibinfo{publisher}{Springer}, \bibinfo{pages}{46--60}.
\newblock
\urldef\tempurl%
\url{https://doi.org/10.1007/978-3-031-14714-2\_4}
\showDOI{\tempurl}
\newblock
\shownote{Free version available at \url{https://arxiv.org/abs/2204.09483}}.


\bibitem[Kudela(2022)]%
        {kudela2022critical}
\bibfield{author}{\bibinfo{person}{Jakub Kudela}.}
  \bibinfo{year}{2022}\natexlab{}.
\newblock \showarticletitle{A critical problem in benchmarking and analysis of
  evolutionary computation methods}.
\newblock \bibinfo{journal}{\emph{Nature Machine Intelligence}}
  \bibinfo{volume}{4}, \bibinfo{number}{12} (\bibinfo{year}{2022}),
  \bibinfo{pages}{1238--1245}.
\newblock


\bibitem[Lacroix and McCall(2019)]%
        {lacroix2019}
\bibfield{author}{\bibinfo{person}{Benjamin Lacroix} {and}
  \bibinfo{person}{John McCall}.} \bibinfo{year}{2019}\natexlab{}.
\newblock \showarticletitle{Limitations of Benchmark Sets and Landscape
  Features for Algorithm Selection and Performance Prediction}. In
  \bibinfo{booktitle}{\emph{Proc. of Genetic and Evolutionary Computation
  (GECCO'19)}} (Prague, Czech Republic). \bibinfo{publisher}{ACM},
  \bibinfo{address}{New York, NY, USA}, \bibinfo{pages}{261–262}.
\newblock
\showISBNx{9781450367486}
\urldef\tempurl%
\url{https://doi.org/10.1145/3319619.3322051}
\showDOI{\tempurl}


\bibitem[Larra{\~n}aga and Lozano(2001)]%
        {emna}
\bibfield{author}{\bibinfo{person}{Pedro Larra{\~n}aga} {and}
  \bibinfo{person}{Jose~A Lozano}.} \bibinfo{year}{2001}\natexlab{}.
\newblock \bibinfo{booktitle}{\emph{Estimation of distribution algorithms: A
  new tool for evolutionary computation}}. Vol.~\bibinfo{volume}{2}.
\newblock \bibinfo{publisher}{Springer Science \& Business Media}.
\newblock


\bibitem[Long et~al\mbox{.}(2023)]%
        {long2023challenges}
\bibfield{author}{\bibinfo{person}{Fu~Xing Long}, \bibinfo{person}{Diederick
  Vermetten}, \bibinfo{person}{Anna~V Kononova}, \bibinfo{person}{Roman
  Kalkreuth}, \bibinfo{person}{Kaifeng Yang}, \bibinfo{person}{Thomas
  B{\"a}ck}, {and} \bibinfo{person}{Niki van Stein}.}
  \bibinfo{year}{2023}\natexlab{}.
\newblock \showarticletitle{Challenges of ELA-guided Function Evolution using
  Genetic Programming}.
\newblock \bibinfo{journal}{\emph{arXiv preprint arXiv:2305.15245}}
  (\bibinfo{year}{2023}).
\newblock


\bibitem[Long et~al\mbox{.}(2022)]%
        {evostar_bbob_instance}
\bibfield{author}{\bibinfo{person}{Fu~Xing Long}, \bibinfo{person}{Diederick
  Vermetten}, \bibinfo{person}{Bas van Stein}, {and} \bibinfo{person}{Anna~V.
  Kononova}.} \bibinfo{year}{2022}\natexlab{}.
\newblock \showarticletitle{{BBOB} Instance Analysis: Landscape Properties and
  Algorithm Performance across Problem Instances}.
\newblock \bibinfo{journal}{\emph{CoRR}}  \bibinfo{volume}{abs/2211.16318}
  (\bibinfo{year}{2022}).
\newblock
\urldef\tempurl%
\url{https://doi.org/10.48550/arXiv.2211.16318}
\showDOI{\tempurl}
\showeprint[arXiv]{2211.16318}


\bibitem[Mersmann et~al\mbox{.}(2011)]%
        {mersmann2011exploratory}
\bibfield{author}{\bibinfo{person}{Olaf Mersmann}, \bibinfo{person}{Bernd
  Bischl}, \bibinfo{person}{Heike Trautmann}, \bibinfo{person}{Mike Preuss},
  \bibinfo{person}{Claus Weihs}, {and} \bibinfo{person}{G{\"u}nter Rudolph}.}
  \bibinfo{year}{2011}\natexlab{}.
\newblock \showarticletitle{Exploratory landscape analysis}. In
  \bibinfo{booktitle}{\emph{Proc. of Genetic and Evolutionary Computation
  (GECCO'11)}}. \bibinfo{publisher}{ACM}, \bibinfo{pages}{829--836}.
\newblock


\bibitem[Mu{\~n}oz et~al\mbox{.}(2022)]%
        {ela2_munoz2022}
\bibfield{author}{\bibinfo{person}{Mario~Andr{\'e}s Mu{\~n}oz},
  \bibinfo{person}{Michael Kirley}, {and} \bibinfo{person}{Kate Smith-Miles}.}
  \bibinfo{year}{2022}\natexlab{}.
\newblock \showarticletitle{Analyzing randomness effects on the reliability of
  exploratory landscape analysis}.
\newblock \bibinfo{journal}{\emph{Natural Computing}} \bibinfo{volume}{21},
  \bibinfo{number}{2} (\bibinfo{year}{2022}), \bibinfo{pages}{131--154}.
\newblock


\bibitem[Mu{\~{n}}oz and Smith{-}Miles(2020)]%
        {NewBBOB-ISA-MunozS20}
\bibfield{author}{\bibinfo{person}{Mario~A. Mu{\~{n}}oz} {and}
  \bibinfo{person}{Kate Smith{-}Miles}.} \bibinfo{year}{2020}\natexlab{}.
\newblock \showarticletitle{Generating New Space-Filling Test Instances for
  Continuous Black-Box Optimization}.
\newblock \bibinfo{journal}{\emph{Evol. Comput.}} \bibinfo{volume}{28},
  \bibinfo{number}{3} (\bibinfo{year}{2020}), \bibinfo{pages}{379--404}.
\newblock
\urldef\tempurl%
\url{https://doi.org/10.1162/evco\_a\_00262}
\showDOI{\tempurl}


\bibitem[Mu{\~n}oz et~al\mbox{.}(2015)]%
        {munoz2015algorithm}
\bibfield{author}{\bibinfo{person}{Mario~A Mu{\~n}oz}, \bibinfo{person}{Yuan
  Sun}, \bibinfo{person}{Michael Kirley}, {and} \bibinfo{person}{Saman~K
  Halgamuge}.} \bibinfo{year}{2015}\natexlab{}.
\newblock \showarticletitle{Algorithm selection for black-box continuous
  optimization problems: A survey on methods and challenges}.
\newblock \bibinfo{journal}{\emph{Information Sciences}}  \bibinfo{volume}{317}
  (\bibinfo{year}{2015}), \bibinfo{pages}{224--245}.
\newblock


\bibitem[Nikolikj et~al\mbox{.}(2023)]%
        {nikolikj2023rf}
\bibfield{author}{\bibinfo{person}{Ana Nikolikj}, \bibinfo{person}{Carola
  Doerr}, {and} \bibinfo{person}{Tome Eftimov}.}
  \bibinfo{year}{2023}\natexlab{}.
\newblock \showarticletitle{RF+ clust for Leave-One-Problem-Out Performance
  Prediction}. In \bibinfo{booktitle}{\emph{Proc. of Applications of
  Evolutionary Computation (Evo Applications'23)}}. Springer,
  \bibinfo{pages}{285--301}.
\newblock


\bibitem[PatrikValkovic(2021)]%
        {BBOBtorch}
\bibfield{author}{\bibinfo{person}{PatrikValkovic}.}
  \bibinfo{year}{2021}\natexlab{}.
\newblock \bibinfo{title}{BBOBtorch}.
\newblock
  \bibinfo{howpublished}{\url{https://github.com/PatrikValkovic/BBOBtorch}}.
\newblock


\bibitem[Pikalov and Mironovich(2022)]%
        {pikalov2022parameter}
\bibfield{author}{\bibinfo{person}{Maxim Pikalov} {and}
  \bibinfo{person}{Vladimir Mironovich}.} \bibinfo{year}{2022}\natexlab{}.
\newblock \showarticletitle{Parameter tuning for the (1+($\lambda$, $\lambda$))
  genetic algorithm using landscape analysis and machine learning}. In
  \bibinfo{booktitle}{\emph{International Conference on the Applications of
  Evolutionary Computation (Part of EvoStar)}}. Springer,
  \bibinfo{pages}{704--720}.
\newblock


\bibitem[Prager(2022)]%
        {pFlacco}
\bibfield{author}{\bibinfo{person}{Raphael~Patrick Prager}.}
  \bibinfo{year}{2022}\natexlab{}.
\newblock \bibinfo{title}{pFlacco}.
\newblock \bibinfo{howpublished}{\url{https://pypi.org/project/pflacco/}}.
\newblock


\bibitem[Rapin and Teytaud(2018)]%
        {nevergrad}
\bibfield{author}{\bibinfo{person}{J{\'{e}}r{\'{e}}my Rapin} {and}
  \bibinfo{person}{Olivier Teytaud}.} \bibinfo{year}{2018}\natexlab{}.
\newblock \bibinfo{title}{{Nevergrad - A gradient-free optimization platform}}.
\newblock
  \bibinfo{howpublished}{\url{https://GitHub.com/FacebookResearch/Nevergrad}}.
\newblock


\bibitem[Renau et~al\mbox{.}(2019)]%
        {Renau2019features}
\bibfield{author}{\bibinfo{person}{Quentin Renau}, \bibinfo{person}{Johann
  Dreo}, \bibinfo{person}{Carola Doerr}, {and} \bibinfo{person}{Benjamin
  Doerr}.} \bibinfo{year}{2019}\natexlab{}.
\newblock \showarticletitle{Expressiveness and Robustness of Landscape
  Features}. In \bibinfo{booktitle}{\emph{Proc. of Genetic and Evolutionary
  Computation (GECCO'19)}} (Prague, Czech Republic). \bibinfo{publisher}{ACM},
  \bibinfo{pages}{2048--2051}.
\newblock
\showISBNx{978-1-4503-6748-6}
\urldef\tempurl%
\url{https://doi.org/10.1145/3319619.3326913}
\showDOI{\tempurl}


\bibitem[Renau et~al\mbox{.}(2021)]%
        {renau2021towards}
\bibfield{author}{\bibinfo{person}{Quentin Renau}, \bibinfo{person}{Johann
  Dr{\'e}o}, \bibinfo{person}{Carola Doerr}, {and} \bibinfo{person}{Benjamin
  Doerr}.} \bibinfo{year}{2021}\natexlab{}.
\newblock \showarticletitle{Towards explainable exploratory landscape analysis:
  extreme feature selection for classifying BBOB functions}. In
  \bibinfo{booktitle}{\emph{Applications of Evolutionary Computation: 24th
  International Conference, EvoApplications 2021, Held as Part of EvoStar 2021,
  Virtual Event, April 7--9, 2021, Proceedings 24}}.
  \bibinfo{publisher}{Springer}, \bibinfo{pages}{17--33}.
\newblock


\bibitem[{\v{S}}kvorc et~al\mbox{.}(2020)]%
        {vskvorc2020understanding}
\bibfield{author}{\bibinfo{person}{Urban {\v{S}}kvorc}, \bibinfo{person}{Tome
  Eftimov}, {and} \bibinfo{person}{Peter Koro{\v{s}}ec}.}
  \bibinfo{year}{2020}\natexlab{}.
\newblock \showarticletitle{Understanding the problem space in single-objective
  numerical optimization using exploratory landscape analysis}.
\newblock \bibinfo{journal}{\emph{Applied Soft Computing}}
  \bibinfo{volume}{90} (\bibinfo{year}{2020}), \bibinfo{pages}{106138}.
\newblock


\bibitem[Smith-Miles and Mu{\~n}oz(2023)]%
        {smith2023instance}
\bibfield{author}{\bibinfo{person}{Kate Smith-Miles} {and}
  \bibinfo{person}{Mario~Andr{\'e}s Mu{\~n}oz}.}
  \bibinfo{year}{2023}\natexlab{}.
\newblock \showarticletitle{Instance space analysis for algorithm testing:
  Methodology and software tools}.
\newblock \bibinfo{journal}{\emph{Comput. Surveys}} \bibinfo{volume}{55},
  \bibinfo{number}{12} (\bibinfo{year}{2023}), \bibinfo{pages}{1--31}.
\newblock


\bibitem[Storn and Price(1997)]%
        {de}
\bibfield{author}{\bibinfo{person}{Rainer Storn} {and} \bibinfo{person}{Kenneth
  Price}.} \bibinfo{year}{1997}\natexlab{}.
\newblock \showarticletitle{Differential evolution-a simple and efficient
  heuristic for global optimization over continuous spaces}.
\newblock \bibinfo{journal}{\emph{Journal of global optimization}}
  \bibinfo{volume}{11}, \bibinfo{number}{4} (\bibinfo{year}{1997}),
  \bibinfo{pages}{341}.
\newblock


\bibitem[Tian et~al\mbox{.}(2020)]%
        {tian2020recommender}
\bibfield{author}{\bibinfo{person}{Ye Tian}, \bibinfo{person}{Shichen Peng},
  \bibinfo{person}{Xingyi Zhang}, \bibinfo{person}{Tobias Rodemann},
  \bibinfo{person}{Kay~Chen Tan}, {and} \bibinfo{person}{Yaochu Jin}.}
  \bibinfo{year}{2020}\natexlab{}.
\newblock \showarticletitle{A recommender system for metaheuristic algorithms
  for continuous optimization based on deep recurrent neural networks}.
\newblock \bibinfo{journal}{\emph{IEEE transactions on artificial
  intelligence}} \bibinfo{volume}{1}, \bibinfo{number}{1}
  (\bibinfo{year}{2020}), \bibinfo{pages}{5--18}.
\newblock


\bibitem[Vermetten et~al\mbox{.}(2023a)]%
        {modDE}
\bibfield{author}{\bibinfo{person}{Diederick Vermetten}, \bibinfo{person}{Fabio
  Caraffini}, \bibinfo{person}{Anna~V Kononova}, {and} \bibinfo{person}{Thomas
  B{\"a}ck}.} \bibinfo{year}{2023}\natexlab{a}.
\newblock \showarticletitle{Modular Differential Evolution}. In
  \bibinfo{booktitle}{\emph{Proc. of Genetic and Evolutionary Computation
  (GECCO'23)}}. \bibinfo{publisher}{ACM}.
\newblock
\urldef\tempurl%
\url{https://doi.org/10.1145/3583131.3590417}
\showDOI{\tempurl}
\newblock
\shownote{To appear. Code available at
  \url{https://github.com/Dvermetten/ModDE}}.


\bibitem[Vermetten et~al\mbox{.}(2023c)]%
        {automl_mabbob}
\bibfield{author}{\bibinfo{person}{Diederick Vermetten},
  \bibinfo{person}{Furong Ye}, \bibinfo{person}{Thomas B{\"a}ck}, {and}
  \bibinfo{person}{Carola Doerr}.} \bibinfo{year}{2023}\natexlab{c}.
\newblock \showarticletitle{{MA}-{BBOB}: Many-Affine Combinations of {BBOB}
  Functions for Evaluating Auto{ML} Approaches in Noiseless Numerical Black-Box
  Optimization Contexts}. In \bibinfo{booktitle}{\emph{AutoML Conference 2023
  (ABCD Track)}}.
\newblock
\urldef\tempurl%
\url{https://openreview.net/forum?id=uN70Dum6pC2}
\showURL{%
\tempurl}


\bibitem[Vermetten et~al\mbox{.}(2023b)]%
        {ABBOB-GECCO}
\bibfield{author}{\bibinfo{person}{Diederick Vermetten},
  \bibinfo{person}{Furong Ye}, {and} \bibinfo{person}{Carola Doerr}.}
  \bibinfo{year}{2023}\natexlab{b}.
\newblock \showarticletitle{Using Affine Combinations of {BBOB} Problems for
  Performance Assessment}. In \bibinfo{booktitle}{\emph{Proc. of Genetic and
  Evolutionary Computation Conference (GECCO'23)}},
  Vol.~\bibinfo{volume}{abs/2303.04573}. \bibinfo{publisher}{ACM}.
\newblock
\urldef\tempurl%
\url{https://doi.org/10.1145/3583131.3590412}
\showDOI{\tempurl}


\bibitem[Vermetten et~al\mbox{.}(2023d)]%
        {zenodo_extension}
\bibfield{author}{\bibinfo{person}{Diederick Vermetten},
  \bibinfo{person}{Furong Ye}, \bibinfo{person}{Carola Doerr}, {and}
  \bibinfo{person}{Thomas Back}.} \bibinfo{year}{2023}\natexlab{d}.
\newblock \bibinfo{title}{{MA-BBOB - Reproducibility and Additional Data}}.
\newblock
\newblock
\urldef\tempurl%
\url{https://doi.org/10.5281/zenodo.10376912}
\showDOI{\tempurl}


\bibitem[Vermetten et~al\mbox{.}(2023e)]%
        {zenodo_automl}
\bibfield{author}{\bibinfo{person}{Diederick Vermetten},
  \bibinfo{person}{Furong Ye}, \bibinfo{person}{Carola Doerr}, {and}
  \bibinfo{person}{Thomas Back}.} \bibinfo{year}{2023}\natexlab{e}.
\newblock \bibinfo{title}{{Many-Affine BBOB Function Combinations - Data and
  Figures}}.
\newblock
\newblock
\urldef\tempurl%
\url{https://doi.org/10.5281/zenodo.7826036}
\showDOI{\tempurl}


\bibitem[Wang et~al\mbox{.}(2022)]%
        {IOHanalyzer}
\bibfield{author}{\bibinfo{person}{Hao Wang}, \bibinfo{person}{Diederick
  Vermetten}, \bibinfo{person}{Furong Ye}, \bibinfo{person}{Carola Doerr},
  {and} \bibinfo{person}{Thomas B{\"{a}}ck}.} \bibinfo{year}{2022}\natexlab{}.
\newblock \showarticletitle{IOHanalyzer: Detailed Performance Analysis for
  Iterative Optimization Heuristic}.
\newblock \bibinfo{journal}{\emph{ACM Trans. Evol. Learn. Optim.}}
  \bibinfo{volume}{2}, \bibinfo{number}{1} (\bibinfo{year}{2022}),
  \bibinfo{pages}{3:1--3:29}.
\newblock
\showISSN{2688-299X}
\urldef\tempurl%
\url{https://doi.org/10.1145/3510426}
\showDOI{\tempurl}
\newblock
\shownote{IOHanalyzer is available at CRAN, on GitHub, and as web-based GUI,
  see \url{https://iohprofiler.github.io/IOHanalyzer/} for links}.


\bibitem[Weise et~al\mbox{.}(2008)]%
        {w_model}
\bibfield{author}{\bibinfo{person}{Thomas Weise}, \bibinfo{person}{Stefan
  Niemczyk}, \bibinfo{person}{Hendrik Skubch}, \bibinfo{person}{Roland
  Reichle}, {and} \bibinfo{person}{Kurt Geihs}.}
  \bibinfo{year}{2008}\natexlab{}.
\newblock \showarticletitle{A tunable model for multi-objective, epistatic,
  rugged, and neutral fitness landscapes}. In
  \bibinfo{booktitle}{\emph{Proceedings of the 10th annual conference on
  Genetic and evolutionary computation}}. \bibinfo{pages}{795--802}.
\newblock


\end{thebibliography}

\end{document}